\documentclass[preprint]{elsarticle}

\usepackage{hyperref}
\usepackage{url}
\usepackage[utf8]{inputenc} 
\usepackage[T1]{fontenc}    
\usepackage{hyperref}       
\usepackage{url}            
\usepackage{booktabs}       
\usepackage{amsfonts}       
\usepackage{nicefrac}       
\usepackage{microtype}      
\usepackage{xcolor}         
\usepackage{graphicx}
\usepackage{mathtools}
\usepackage{amsmath}
\usepackage{amssymb}
\usepackage{color, colortbl}
\usepackage{cancel}
\usepackage{caption}
\usepackage{subcaption}
\usepackage{algpseudocode}
\usepackage{algorithm}
\usepackage{multirow}
\usepackage{xfrac}
\usepackage{lineno}
\usepackage{algpseudocode}
\usepackage{algorithm}
\usepackage{array}
\usepackage{fixltx2e}
\usepackage{stfloats}
\usepackage{picins}
\allowdisplaybreaks

\usepackage[font=footnotesize,labelfont=bf]{caption}
\usepackage[font=footnotesize,labelfont=bf]{subcaption}

\journal{Pattern Recognition}

\begin{document}

\begin{frontmatter}

\title{EdVAE: Mitigating Codebook Collapse with Evidential Discrete Variational Autoencoders}


\author[1,2]{Gulcin Baykal\corref{cor1}}
\cortext[cor1]{Corresponding author}
\ead{baykalg@itu.edu.tr}

\author[2]{Melih Kandemir}
\ead{kandemir@imada.sdu.dk}

\author[3]{Gozde Unal}
\ead{gozde.unal@itu.edu.tr}

\affiliation[1]{organization={Department of Computer Engineering, 
                              Istanbul Technical University},
                city={Istanbul},
                country={Türkiye}}

\affiliation[2]{organization={Department of Mathematics and Computer Science,
                              University of Southern Denmark},
                city={Odense},
                country={Denmark}}

\affiliation[3]{organization={Department of AI and Data Engineering, 
                              Istanbul Technical University},
                city={Istanbul},
                country={Türkiye}}

\begin{abstract}
    Codebook collapse is a common problem in training deep generative models with discrete representation spaces like Vector Quantized Variational Autoencoders (VQ-VAEs). We observe that the same problem arises for the alternatively designed discrete variational autoencoders (dVAEs) whose encoder directly learns a distribution over the codebook embeddings to represent the data. We hypothesize that using the softmax function to obtain a probability distribution causes the codebook collapse by assigning overconfident probabilities to the best matching codebook elements. In this paper, we propose a novel way to incorporate evidential deep learning (EDL) through a hierarchical Bayesian modeling instead of softmax to combat the codebook collapse problem of dVAE. We evidentially monitor the significance of attaining the probability distribution over the codebook embeddings, in contrast to softmax usage. Our experiments using various datasets show that our model, called EdVAE, mitigates codebook collapse while improving the reconstruction performance, and enhances the codebook usage compared to dVAE and VQ-VAE based models. Our code can be found at \url{https://github.com/ituvisionlab/EdVAE}.
\end{abstract}

\begin{keyword}
Vector Quantized Variational Autoencoders, Discrete Variational Autoencoders, Evidential Deep Learning, Codebook Collapse
\end{keyword}

\end{frontmatter}

\section{Introduction}\label{sec:intro}

Generative modeling of images has been one of the popular research themes that aided in advancement of deep neural networks, particularly in enhancement of unsupervised learning models like Variational Autoencoders (VAEs) \citep{kingma2013auto}, Generative Adversarial Networks (GANs) \citep{goodfellow2014generative}, and Diffusion models \citep{ho2020denoising}. VAEs \citep{kingma2013auto} and their variant models have shown promise as solutions to numerous problems in generative modelling such as disentanglement of the representations \citep{higgins2017beta}, discretization of the representations \citep{oord2017neural, ramesh2021zeroshot}, and high-quality image generation \citep{vahdat2020nvae, child2020verydv}. Although most of the VAE models assume a continuous latent space to represent the data, discrete representations are more suitable to express categories that modulate the observation space \citep{oord2017neural}. Supporting this rationale, recent celebrated large generative models like VQGAN \citep{esser2020taming}, LDM \citep{rombach2021highresolutionis}, and DALL-E 1 \citep{ramesh2021zeroshot} also rely on discrete latent spaces learned by discrete VAEs to describe the image data. 

It is customary to form the latent space as a \textit{codebook} consisting of discrete latent embeddings, where those embeddings are learned to represent the data. VQ-VAE \citep{oord2017neural,razavi2019generating} and its variants \citep{esser2020taming,SUN2024109962} are discrete VAEs that quantize the encoded representation of the data by an encoder with the nearest latent embedding in the learnable codebook in a deterministic way. VQ-VAEs achieve considerably high reconstruction and generation performances. However, they are observed to suffer from the \textit{codebook collapse} problem defined as the under-usage of the codebook embeddings, causing a redundancy in the codebook and limiting the expressive power of the generative model. As the deterministic quantization is the most likely cause of the codebook collapse problem in VQ-VAEs \citep{takida2022sqvae}, probabilistic approaches \citep{takida2022sqvae}, optimization changes \citep{huh2023improvedvqste} as well as codebook reset \citep{williams2020hierarchicalqa, dhariwal2020jukebox} and hyperparameter tuning \citep{dhariwal2020jukebox} are employed in VQ-VAEs to combat the codebook collapse problem. 

Unlike VQ-VAEs, the encoder of another discrete VAE, which is called dVAE \citep{ramesh2021zeroshot}, learns a \textit{distribution} over the codebook embeddings for each latent in the representation. That means, instead of quantizing a latent with a single, deterministically selected codebook embedding, the encoder of dVAE incorporates stochasticity to the selection of the embeddings where the learned distribution is modeled as a \textit{Categorical} distribution. We find out that dVAE also suffers from the codebook collapse problem even though stochasticity is involved. One of our hypotheses is that attaining the probability masses for codebook embeddings using a softmax function induces a codebook collapse in dVAE, as we demonstrate in this paper.

Softmax notoriously overestimates the probability mass of the prediction, which in turn led to exploration of different alternatives to softmax, especially in  classification tasks \citep{joo2020beingba,sensoy2018evidential}. Among those approaches, the widely adopted EDL \citep{sensoy2018evidential} places a Dirichlet distribution whose concentration parameters over the class probabilities are learned by the encoder. In EDL, the class predictions are considered as the subjective opinions, and the evidences leading to those opinions are collected from the data, which are explicitly used as the concentration parameters of the Dirichlet distribution. To define such a framework, the softmax layer of the encoder is removed, and the logits of the encoder are used as the concentration parameters whose mean values are used as the predicted class probabilities. EDL can be also viewed as a generative model where the class labels follow a normal distribution whose mean is set by the uninformative Dirichlet prior over the class probabilities \citep{kandemir2022evidential}.

In this work, we find out that the root cause of the codebook collapse in dVAE can be framed as the artificial intelligence counterpart of the cognitive psychological phenomenon called the \textit{confirmation bias} \citep{kahneman2011thinking}. This bias is developed cumulatively during the whole training process as the model overconfidently relates new observations to those already seen ones. We demonstrate by way of experiments that the spiky softmax probabilities lead to the confirmation bias problem. In order to mitigate the latter, we introduce an uncertainty-aware mechanism to map the inputs to the codebook embeddings by virtue of an evidential formulation. To that end, dVAE encoder collects evidences from the data, and the codebook embeddings that represent the data are selected based on those evidences. While the highest evidence increases the probability of the corresponding embedding to be selected, the collected evidences lead to a smoother probability distribution compared to softmax probabilities which leads to a diversified codebook usage. We reformulate the original EDL framework, and build a hierarchical Bayesian extension of dVAE. We summarize our contributions as follows: (1) We introduce an original extension of dVAE that is a hierarchical Bayesian model using Dirichlet-Categorical distributions by virtue of EDL incorporation. (2) We report evidence of the confirmation bias problem caused by the softmax probabilities used in dVAE. (3) We observe that the EDL integration improves the codebook usage of dVAE, which is measured by the perplexity.

In our work, we set the baseline results of dVAE for various datasets, and surpass the baseline measures with our model called Evidential dVAE (EdVAE) in terms of reconstruction performance, codebook usage, and image generation performance in most of the settings. We also compare EdVAE with state-of-the-art VQ-VAE based models using various experimental settings to demonstrate that our model performs close to or better than the VQ-VAE based methods.

\section{Related Work}\label{sec:rel_work}

\textbf{Codebook Collapse on VQ-VAEs:} Vector quantization, which is useful for various tasks including image compression \citep{theis2017lossy, agustsson2017soft}, is the backbone of the VQ-VAE \citep{oord2017neural}. While deterministically trained VQ-VAEs show favorable performance on image reconstruction and generation \citep{razavi2019generating, esser2020taming}, text decoding \citep{kaiser18fast}, music generation \citep{dhariwal2020jukebox}, and motion generation \citep{siyao2022bailando}, some of the VQ-VAE variants use stochasticity and other tricks during the training, especially to alleviate the codebook collapse problem.

Codebook reset trick \citep{williams2020hierarchicalqa} replaces the unused codebook embeddings with the perturbed version of the most used codebook embedding during the training, and increases the number of embeddings used from the codebook. New perspectives on comprehending VQ-VAEs such as affine re-parameterization of the codebook embeddings, alternated optimization during the training, and synchronized update rule for the quantized representation are proposed by \citet{huh2023improvedvqste} to address the problems of VQ-VAEs including the codebook collapse.  

To incorporate stochasticity, a soft expectation maximization (EM) algorithm is reformulated based on the hard EM modeling of the vector quantization \citep{roy2018theory}. GS-VQ-VAE \citep{Snderby2017ContinuousRT} uses the Euclidean distance between the encoder's output and the codebook as the parameters of a Categorical distribution, and the codebook embeddings are selected by sampling. SQ-VAE \citep{takida2022sqvae} defines stochastic quantization and dequantization processes which are parameterized by probability distributions. Those stochastic processes enable codebook usage implicitly without needing additional tricks such as codebook resetting.

Although the codebook collapse problem of the VQ-VAEs is studied in detail with observed success, it is still an open question for dVAEs. The dVAEs are employed instead of VQ-VAEs in \citep{ramesh2021zeroshot} to obtain an image representation for the text-to-image generation problem. As the dVAEs have shown great potential in such complicated tasks, specifying existing problems of dVAE and providing relevant solutions are poised to bring both methodological and practical benefits. 

To that end, our work proposes a novel way to combat the codebook collapse problem of the dVAEs. While the proposed methods for VQ-VAEs have the same objective of mitigating the codebook collapse problem like our method, our intuition differs since we focus on the properties of the distribution learned over the codebook, and try to attain a better distribution. On the other hand, other methods focus on the internal dynamics of the VQ-VAE model and its training.

\textbf{Hierarchical Bayesian Models:} Hierarchical Bayesian models have diverse variants like belief networks \citep{pearl2009probabilistic}, Bayesian neural networks \citep{neal2012bayesian}, applied in healthcare, finance, and machine learning. Hierarchical VAEs are the realization of hierarchical Bayesian modeling in the context of deep generative models. Some hierarchical VAEs, exemplified by models like the Ladder VAE \citep{ladder_vae}, leverage a layered structure to capture hierarchical dependencies in data. In contrast, models like VQ-VAE-2 \citep{razavi2019generating} adopt a different approach to hierarchical VAEs. VQ-VAE-2 focuses on discrete latent codes organized hierarchically within a codebook. However, our model exhibits a hierarchical structure by modeling uncertainty in the parameters of the Categorical distribution, and it involves sampling latent variables from these distributions.

\textbf{Evidential Deep Learning:} The foundational work by \citet{sensoy2018evidential} has been pivotal in advancing uncertainty quantification in deep learning, and its methodology for modeling uncertainty  inspires subsequent research. \citet{edl_molecular} applies EDL to molecular property prediction, \citet{edl_openset} explores its use in open-set action recognition, and \citet{edl_stereo} focuses on uncertainty estimation for stereo matching. These works collectively demonstrate the versatility and impact of the EDL framework, showcasing its influence across diverse domains. In our work, we employ EDL within a VAE-based framework for the first time. This innovative incorporation leads to a hierarchical Bayesian model using Dirichlet-Categorical distributions.

\section{Background}\label{sec:background}

\subsection{Discrete Variational Autoencoders}\label{sec:dvae}
Discrete VAEs aim to model the high-dimensional data $x$ with the low-dimensional and discrete latent representation $z$ by maximizing the ELBO of the log-likelihood of the data:
\begin{equation}
    \mathcal{L}_{\text{ELBO}} = \mathbb{E}_{q(z|x)}[\log p(x|z)] - \text{KL}[q(z|x) || p(z)] .
\label{eq:elbo}
\end{equation}
where $p(x|z)$ is the generative model which is designed as a decoder, $q(z|x)$ is the approximated posterior which is designed as an encoder, and $p(z)$ is a prior. 

In discrete VAEs, $d$ dimensional $z$ is sampled from a Categorical distribution over the possible values of $z$ as $z \sim Cat(z|\pi)$. Encoder returns unnormalized log probabilities $l=[l_1, ..., l_d]$ and $\pi=softmax(l)$ is used to obtain the probability masses of this Categorical distribution. Then, $z$ is fed into the decoder to reconstruct $x$.

As the discrete variables sampled from a Categorical distribution do not permit an end-to-end training, methods for continuous relaxation of the discrete variables are proposed, introducing the Gumbel-Softmax trick \citep{jang2016categorical, maddison2016concrete}. Gumbel-Softmax trick briefly introduces Gumbel noises $g=[g_1, ..., g_d], g_i \sim Gumbel(0, 1)$ into $l$ to introduce randomness so that:
\begin{equation}    
    z = argmax_d(l+g) \sim Cat(z|\pi),
    \label{eq: gumbel_max}
\end{equation}
holds. While Equation~\ref{eq: gumbel_max} is the first step to take a differentiable sample from a Categorical distribution, taking the $argmax$ is also non-differentiable. Therefore, softmax operation with a temperature parameter $\tau$:
\begin{equation}
    z_m = \frac{e^{\frac{l_m+g_m}{\tau}}}{\sum_{j=1}^{k}e^{\frac{l_j+g_j}{\tau}}},
    \label{eq: gumbel_max_2}
\end{equation}
where $m \in \{1, \dots , d\}$ can be used to approximate $argmax$ operation, and $\tau$ controls how closely the Gumbel-Softmax distribution approximates the Categorical distribution, resulting in a differentiable $z$.

VQ-VAE is a discrete VAE variant where a learnable codebook $\mathcal{M} \in \mathbb{R}^{K \times D}$ consisting $K$ number of $D$ dimensional embeddings is trained to represent a dataset. In VQ-VAE, discrete latent variables are used to retrieve embeddings from $\mathcal{M}$ so that these embeddings can represent $x$. These discrete latent variables are obtained as follows: an encoder constructs a continuous latent representation $z_e(x) \in \mathbb{R}^{N \times N \times D}$ spanned by $N \times N$ matrices, and the discrete latent $z \in \mathbb{R}^{N \times N \times K}$ is \textit{deterministically} obtained using the one-hot representations of the closest embeddings' indices $\mathcal{M}$ for each $N \times N$ spatial location using Euclidean distance as follows:
\[ q(z_i=\text{one-hot(k)}|x) =
  \begin{cases}
    1       & \quad \text{if } k=\text{argmin}_{j}||z_e(x)_i-e_j||_2 \\
    0  & \quad \text{otherwise} 
  \end{cases}
\]
where $i \in \{1, \dots, N \times N\}$, $k \in \{1, \dots, K\}$, $z_e(x)_i \in \mathbb{R}^D$ is the $i^{th}$ vector in $z_e(x)$, and $e_j \in \mathbb{R}^D$ is a codebook embedding. Then, a quantized representation $z_q(x)=z*\mathcal{M}$ is obtained using the indices $z$s to retrieve the corresponding codebook embeddings. The operator * performs tensor-matrix multiplication where each $z_i$ for each spatial position will retrieve the related codebook vectors from codebook matrix $\mathcal{M}$. Lastly, the quantized representation $z_q(x) \in \mathbb{R}^{N \times N \times D}$ is fed into the decoder to reconstruct $x$. The value of $N$ changes depending on the downsampling amount of the encoder (see Appendix~\ref{appendix:experimental_details} for encoder design).

Unlike VQ-VAEs, dVAE \citep{ramesh2021zeroshot} compresses $x$ into $z_e(x) \in \mathbb{R}^{N \times N \times K}$ which is taken as the unnormalized log probabilities of a Categorical distribution over $K$ different codebook embeddings for each $N \times N$ spatial location. Then, the quantizing discrete codebook indices are sampled as follows:
\begin{equation}
    \pi = softmax(z_e(x)), \qquad z \sim Cat(z|\pi), \qquad z_q(x) = z*\mathcal{M}.
    \label{eq:dvae_eq}
\end{equation}
While VQ-VAE obtains the discrete latent $z$ deterministically, dVAE samples them as in Equation~\ref{eq:dvae_eq}. dVAE also incorporates Gumbel-Softmax relaxation for differentiability of the sampled discrete variables $z$s in Equation~\ref{eq: gumbel_max} and Equation~\ref{eq: gumbel_max_2}.

\subsection{Hierarchical Bayesian Models}\label{sec:hbm}
A Hierarchical Bayesian model is defined as a chain of random variables following a hierarchical structure, where each level represents a different level of abstraction or variability in the data. Given a data set $X=\{x_i | i=1,\ldots,L\}$ of $L$ observations, assume the generation of this data set is governed by two groups of local latent variables $U$ and $Z$ that are independent across data points. The joint distribution reads:
\begin{align*}
    p(X,Z,U) &= p(X|Z,U) p(Z|U) p(U) = \prod_i^L p(x_i|z_i,u_i) p(z_i|u_i) p(u_i),
\end{align*}
where the first equality follows from the assumed hierarchy starting from the root node $U$ going down through $Z$ to $X$, and the second from the independence assumption of the latent variables. The resulting data generating process can then be written as:
\begin{equation}
  u_i \sim p(u_i), \qquad z_i | u_i \sim p(z_i | u_i), \qquad x_i | z_i, u_i \sim p(x_i|z_i,u_i). \nonumber
\end{equation}
$z_i$ could be a high level conceptual representation of $x_i$, e.g. determinant features of an object and $u_i$ an even higher level representation, such as properties of object groups. The fact that these variables are linked via probability distributions also accounts for random factors in a way that accumulates downward in the hierarchy.

A hierarchical deep generative network represents the conditional relationships in this data generating process as Neural Networks (NNs). For instance, Hierarchical VAEs do that by introducing hierarchical layers of latent variables and mapping the conditional distributions $p(x_i|z_i,u_i), p(z_i|u_i)$, and $p(u_i)$ to the parameters of NNs. Specifically, the encoder network maps the observed data $x_i$ to latent variables $z_i$ and $u_i$, while the decoder network reconstructs the data from sampled latent variables. The hierarchical structure facilitates the learning of increasingly abstract representations of the data.

\subsection{Evidential Deep Learning as a Hierarchical Bayesian Model}\label{sec:edl}
EDL enhances the NNs in order to allow for probabilistic uncertainty quantification for various tasks such as classification \citep{sensoy2018evidential} or regression \citep{amini2020deep}. As the deterministic NNs output a point estimate for the given input which does not comprise uncertainty over the decision, the EDL changes this behaviour such that the output of the NN is a set of probabilities that represent the likelihood of each possible outcome, by using the evidential logic. EDL implements a hierarchical Bayesian model for a feed-forward classification task as:
\begin{equation}
    u_i|x_i \sim p(u_i|x_i), \qquad z_i|u_i \sim p(z_i|u_i), \qquad y_i|z_i  \sim p(y_i|z_i)
    \label{eq:edl}
\end{equation}
where the hyperprior $u_i$ is conditioned on an input observation $x_i$, and $y_i$ is the class label of $x_i$. EDL's loss function can be cast as a variational inference problem where $q(u, z|x) = p(z|u)p(u|x)$.

\section{Method}\label{sec:method}

\begin{figure}[t!]
    \centering
    \includegraphics[width=0.7\textwidth]{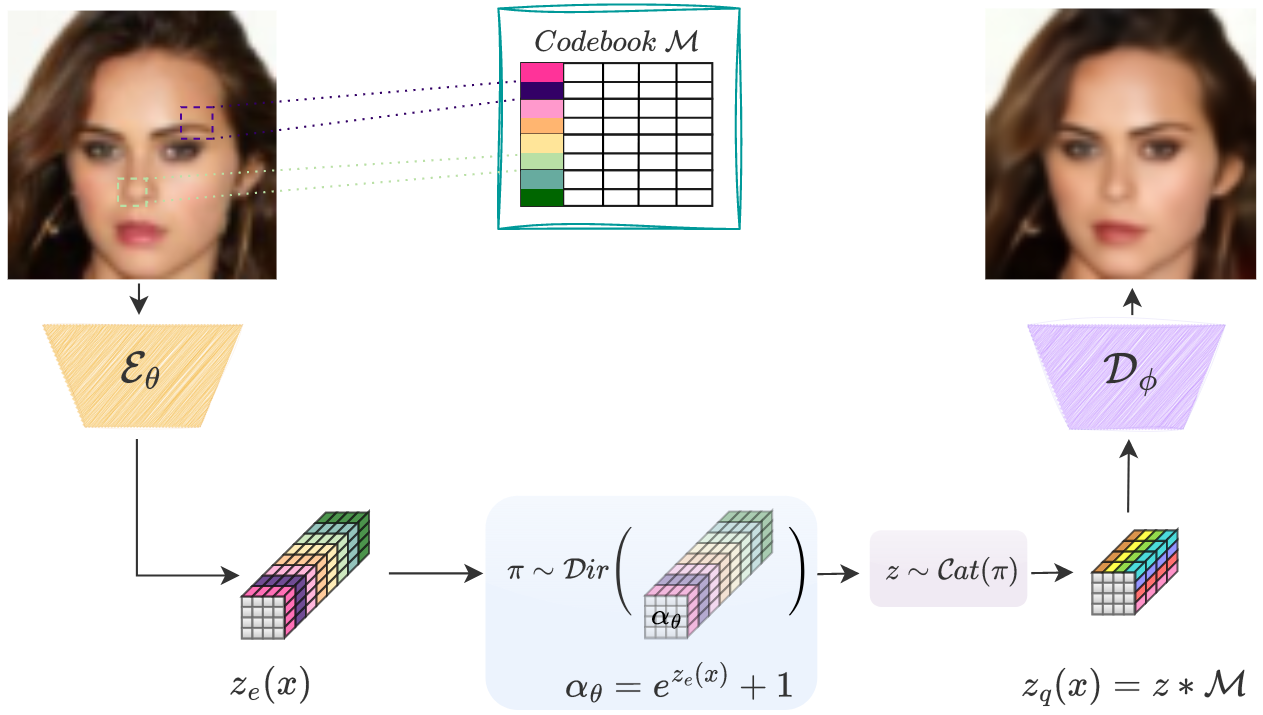}
    \caption{Overview of the method. An illustrative codebook is defined as $\mathcal{M} \in R^{8\times 4}$ where 8 is the number of the codebook embeddings, 4 is the dimensionality of each embedding. For each 16 spatial positions in $z_e(x)$ where $N$ is 4, we define a Dirichlet prior over the parameters of the Categorical distributions which models the codebook embedding assignment to each spatial position.}
    \label{fig:method}
\end{figure}

In this work, we enhance the dVAE model with an evidential perspective to mitigate the codebook collapse problem. To establish a stronger connection between the motivation and the proposed approach, it is essential to delve deeper into the root cause of codebook collapse. This issue arises when the model consistently relies on previously used codebook embeddings throughout training. Conceptually, the phenomenon of confirmation bias aligns with the codebook collapse problem. Over the course of training, the model gradually develops a bias, excessively associating new observations with previously encountered ones due to \textbf{overconfidence}. Given that the softmax function has been acknowledged to exhibit overconfidence issues \citep{joo2020beingba, sensoy2018evidential}, our motivation stems from the need to address the overconfidence challenge. Leveraging EDL, a framework demonstrated to alleviate the overconfidence problem associated with the softmax function, becomes a compelling solution. To sum up, overconfidence is the key connection between the evidential perspective and the codebook collapse problem, and EdVAE strategically leverages EDL's capabilities to counteract the codebook collapse.

Overview of the proposed method is shown in Figure~\ref{fig:method}. The encoder $\mathcal{E}_\theta$'s continuous latent space $z_e(x) \in R^{N\times N\times K}$ spanned by $N \times N$ matrices as explained in Section~\ref{sec:dvae} is in agreement with the goal of learning a distribution over $K$ number of embeddings. Originally, $z_e(x)$ in Figure~\ref{fig:method} is passed to a softmax activation to obtain the parameters $\pi$s of the Categorical distribution that models the codebook embedding assignment in dVAE as explained in Section~\ref{sec:dvae}. In EdVAE, we define a higher-level Dirichlet prior over $\pi$s, and treat $z_e(x)$ as the concentration parameters $\alpha_\theta$ to be learned of this distribution. Therefore, EDL approach builds a hierarchical Bayesian model of dVAE. In Figure~\ref{fig:method}, the areas highlighted by blue and gray show the hierarchy between the distributions. Unlike dVAE which has a single level of Categorical distribution as highlighted by gray, we employ a hierarchical level in EdVAE highlighted by blue, in order to incorporate randomness with well quantified uncertainty over the selection process of the codebook embeddings by virtue of Dirichlet prior. 

EDL incorporation models a second-order uncertainty in EdVAE such that $\alpha_\theta$ represents how confident this prediction is, and $\pi$ predicts which codebook element can best reconstruct the input. The incorporated uncertainty awareness is expected to increase the codebook usage in EdVAE which is highly limited in dVAE due to the softmax operation. The intuition behind this expectation originates from the definition of the codebook collapse. New codebook elements are employed whenever the existing ones cannot explain the newcoming observation. Codebook collapse occurs when the model's prediction is wrong on whether the learned representations are capable of reconstructing the new observation or not. Therefore, employing uncertainty awareness over the codebook embedding selection enables the model to use unused codebook embeddings when it is uncertain about the codebook embedding selection. To validate this, we conduct an experiment revealing a correlation between uncertainty values and perplexity in CIFAR10, and this correlation supports our intuition (see Section~\ref{subsec:uncertainty_effect}).

\subsection{EdVAE Design}
In order to incorporate an evidential mechanism into the dVAE model, the forward model and the design choices should be properly arranged. When we design our forward model, we follow a similar form to the latent variable modeling of EDL as described in Section~\ref{sec:edl} where $p(u_i|x)$ in Equation~\ref{eq:edl} is modeled as a Dirichlet distribution over the Categorical distribution $p(z_i|u_i)$. The concentration parameters, $\alpha_\theta$, of the Dirichlet distribution are defined to be greater than or equal to 1. Therefore, $z_e(x)$ is passed through an exp(.) operation to obtain \textit{evidences}, and 1 is added to obtain the concentration parameters as follows:
\begin{equation}
    \alpha_\theta(x) = \text{exp}(z_e(x)) + 1 .
    \label{eq:alpha}
\end{equation}
We define our forward model to be:
\begin{align}
    p(\pi) &= \mathcal{D}\textit{ir}(\pi | 1,\ldots, 1), \label{eq:prior}\\
    Pr(z|\pi) &=\mathcal{C}\textit{at}(z|\pi), \label{eq:categorical}\\
    p(x|\mathcal{M},z=k) &=\mathcal{N}(x|\mathcal{D}_\phi(\mathcal{M}, z),\sigma^2 I). \label{eq:likelihood}
\end{align}
In dVAE, the prior is defined as a uniform distribution over the codebook embeddings. Equation~\ref{eq:prior} demonstrates our prior design as a Dirichlet distribution that generates uniform distributions over the codebook embeddings on average, and $\pi = [\pi_1, \ldots, \pi_K]$. Equation~\ref{eq:categorical} shows that we model the embedding selection as a Categorical distribution where $z$ is the index of the sampled codebook embedding from $\mathcal{M}$, and the parameters $\pi$s of the Categorical distribution are sampled from the Dirichlet prior. During the training, we obtain samples from the Categorical distribution using Gumbel-Softmax relaxation to backpropagate gradients to the encoder. We decay the temperature parameter of the Gumbel-Softmax to 0 as described in \citep{jang2016categorical} so that the soft quantization operation turns into the hard quantization. Therefore, there may be multiple embedding dimensions that are chosen independently due to the temperature value at train-time. At test-time, we perform hard quantization and simply take a single sample from Equation~\ref{eq:categorical}. The algorithms of the training and the inference of EdVAE are given in Algorithm~\ref{alg:train} and Algorithm~\ref{alg:inference}, respectively. $t$ denotes the index of the training iterations, and $b$ denotes the batch index in the inference. RelaxedOneHotCategorical(.) distribution is differentiable, and the samples from this distribution are soft one-hot vectors. On the other hand, Categorical(.) distribution is not differentiable which is not required during the inference. The samples from this distribution are hard one-hot vectors which indicate one-to-one quantizations.

\begin{algorithm}[t!]
\caption{Training algorithm of EdVAE}\label{alg:train}
\begin{algorithmic}
\State {\bfseries Input:} Dataset $\textbf{x}_{\text{train}}$
\State {\bfseries Output:} Reconstructed $\textbf{x}_{\text{train}}$
\State Initialize the encoder $\mathcal{E}^{[0]}_\theta$, the decoder $\mathcal{D}^{[0]}_\phi$, the codebook $\mathcal{M}^{[0]}$, 
\State and the temperature parameter $\tau^{[0]}=1.0$
\For{$t=1,2,\ldots,T$}
\State $x \leftarrow$ Random minibatch from $\textbf{x}_{\text{train}}$
\State $z_e(x) \leftarrow \mathcal{E}^{[t-1]}_\theta(x)$
\State $\alpha_\theta \leftarrow e^{z_e(x)} + 1$
\State $\pi \sim \mathcal{D}\textit{ir}(\pi|\alpha_\theta)$
\State $z \sim \text{RelaxedOneHotCategorical}(\text{temperature}=\tau^{[t-1]}, \text{probs}=\pi)$
\State $\hat{x} \leftarrow \mathcal{D}^{[t-1]}_\phi(\mathcal{M}, z)$
\State $g \leftarrow \nabla_{\mathcal{M}, \theta, \phi}\mathcal{L}(\mathcal{M}^{[t-1]}, \theta^{[t-1]}, \phi^{[t-1]})$
\State \hspace{60pt} with sampled $x$ and $\hat{x}$
\State $\mathcal{M}^{[t]}, \theta^{[t]}, \phi^{[t]} \leftarrow \text{Update parameters using}$ $g$
\State $\tau^{[t]} \leftarrow \text{CosineAnneal}(\tau^{[t-1]}, t)$
\EndFor
\end{algorithmic}
\end{algorithm}

\begin{algorithm}[t!]
\caption{Inference algorithm of EdVAE}\label{alg:inference}
\begin{algorithmic}
\State {\bfseries Input:} Dataset $\textbf{x}_{\text{test}}$
\State {\bfseries Output:} Reconstructed $\textbf{x}_{\text{test}}$
\State Freeze the parameters of the trained encoder $\mathcal{E}_\theta$, the trained decoder $\mathcal{D}_\phi$, 
\State and the trained codebook $\mathcal{M}$
\For{$b=1,2,\ldots,B$}
\State $x_b \leftarrow$ Minibatch from $\textbf{x}_{\text{test}}$
\State $z_e(x) \leftarrow \mathcal{E}_\theta(x)$
\State $\alpha_\theta \leftarrow e^{z_e(x)} + 1$
\State $\pi \sim \mathcal{D}\textit{ir}(\pi|\alpha_\theta)$
\State $z \sim \text{Categorical}(\text{probs}=\pi)$
\State $\hat{x} \leftarrow \mathcal{D}_\phi(\mathcal{M}, z)$
\EndFor
\end{algorithmic}
\end{algorithm}

In Equation~\ref{eq:likelihood}, we define our likelihood $p(x|\mathcal{M},z=k)$ as a Normal distribution, and $\mathcal{D}_\phi$ denotes the decoder network. The input of the decoder can be formed $z_q(x) = z * \mathcal{M}$ as explained in Section~\ref{sec:dvae}.

We approximate the intractable true posterior $p(\pi,z|x)$ by structured variational inference. Thanks to the Dirichlet-Categorical conjugacy, we use the following approximate distribution that accounts for the full factorization of the codebook elements and the uncertainty on their probabilities:
\begin{align}
    q(\pi, z|x) = \mathcal{C}\textit{at}(z|\pi) \mathcal{D}\textit{ir}(\pi | \alpha_\theta^1(x), \ldots, \alpha_\theta^K(x)).
    \label{eq:posterior}
\end{align}
where $\alpha_\theta(x) = \left [\alpha_\theta^1(x), \ldots, \alpha_\theta^K(x) \right ]$. 
Our hierarchical Bayesian model has a Normal distributed likelihood function the mean of which takes codebook elements chosen by the Categorical distribution as the input and maps them to the image space. It follows the Bayesian structure explained in Section~\ref{sec:hbm}. Even though the Dirichlet-Categorical conjugacy does not grant us an analytical solution for the posterior distribution, it simplifies the computations of the ELBO detailed in Appendix~\ref{appendix:derivations}.

As a result of our derivations, ELBO to be maximized during the training is:
\begin{equation}
    \mathcal{L}(\mathcal{M}, \theta, \phi) = \mathbb{E}_{Pr(z |\pi)} \left [ \mathbb{E}_{q(\pi|x)} [ \log p(x|\mathcal{M},z)] \right ]  - \mathcal{D}_{\text{KL}}(q(\pi|x) || p(\pi)). \label{eq:loss}
\end{equation}
The second term in Equation~\ref{eq:loss} is the Kullback-Leibler divergence between two Dirichlet distributions, hence has the analytical solution as derived in Appendix~\ref{appendix:kl_derivation}. After further derivations given in Appendix~\ref{appendix:derivations} over the first term in Equation~\ref{eq:loss}, our loss function is derived in Equation~\ref{eq:overall_loss}:
\begin{equation}
    \mathcal{L}(\mathcal{M}, \theta, \phi) = \mathbb{E}_{q(\pi|x)} \left [(x-\mathcal{D}_{\phi}(\mathcal{M}, z))^2 \right ] - \beta\mathcal{D}_{\text{KL}}(\mathcal{D}\textit{ir}(\pi|\alpha_\theta(x)) || \mathcal{D}\textit{ir}(\pi|1, ..., 1))
    \label{eq:overall_loss}
\end{equation}
which is optimized to increase the likelihood, while decreasing the KL distance regularized with $\beta$ coefficient \citep{higgins2017beta} between the amortized posterior and the prior. We can perform the optimization in an end-to-end manner as we turn the only non-differentiable part in our hierarchical model that is sampling from a Categorical distribution into a differentiable operation RelaxedOneHotCategorical with Gumbel-Softmax as explained in Section~\ref{sec:dvae} and shown in Algorithm~\ref{alg:train}. Therefore, we use Adam optimizer with an initial learning rate $1e^{-3}$ to optimize our model.

The first term in Equation~\ref{eq:overall_loss} indicates the reconstruction error in terms of mean squared error (MSE) between the input and the reconstruction. The objective goal of the model is to obtain a lower reconstruction error presuming that a well representing latent space is constructed. The second term indicates that the Dirichlet distribution which defines a distribution over the Categorical distributions should converge to a generated distribution in a uniform shape on average in order to represent the codebook embedding selection in a diverse way. When the distribution over the codebook embeddings is uniform, codebook usage is maximized as the probabilities of each codebook embedding to be selected become similar in value. Therefore, while the first term in Equation~\ref{eq:overall_loss} aids directly in building the representation capacity of the latent space, the second term in Equation~\ref{eq:overall_loss} indirectly supports that goal via a  diversified codebook usage. We use a $\beta$ coefficient to balance the reconstruction performance and the codebook usage.

EdVAE introduces a hierarchical structure by learning a Dirichlet distribution over the parameters of a Categorical distribution, with latent variables sampled from this distribution. This approach suits the hierarchical Bayesian modeling basics we present in Section~\ref{sec:hbm} while it differs from other hierarchical VAEs in several aspects. Firstly, the hierarchy is expressed through probabilistic modeling of the parameters, rather than through discrete latent codes or layered representations, in order to increase the codebook entropy. Secondly, while the model captures uncertainty in the parameters, it does not explicitly learn hierarchical representations of the data. Lastly, variational inference involves approximating the posterior distribution over latent variables through the learned Dirichlet-Categorical relationship. Overall, EdVAE introduces a unique approach to hierarchical VAEs, emphasizing probabilistic modeling of parameters and differing from others in terms of representation learning and variational inference strategies.
\section{Experiments}\label{sec:experiments}

\subsection{Experimental Settings}

We perform experiments on CIFAR10 \citep{Krizhevsky09}, CelebA \citep{liu2015deep}, and LSUN Church \citep{yu15lsun} datasets to demonstrate the performance of EdVAE compared to the baseline dVAE and VQ-VAE based methods. We use VQ-VAE-EMA which updates the codebook embeddings with exponential moving averages, and GS-VQ-VAE as the basic VQ-VAE models along with the state-of-the-art VQ-VAE based methods including SQ-VAE and VQ-STE++ mitigating codebook collapse problem. We use the same architectures and hyperparameters described in the original papers and official implementations for a fair comparison. We repeat all of our experiments using three different seeds. Hyperparameter choices and architectural designs are further detailed in Appendix~\ref{appendix:experimental_details}.

\subsection{Evaluations}
\subsubsection{Effects of the softmax distribution}\label{subsec:softmax_effect}

Our hypothesis is that the spiky softmax distribution over the codebook embeddings develops confirmation bias, and the confirmation bias causes a codebook collapse. We test our hypothesis by comparing the average entropy of the probability distributions learned by the encoders of dVAE and EdVAE during the training using CIFAR10 dataset. Low entropy indicates a spiky distribution while high entropy represents a distribution closer to a uniform shape, which is the desired case for a diverse codebook usage. A spiky probability distribution yields a confirmation bias because the codebook embedding with the highest probability mass is favorably selected. On the other hand, when the Categorical distribution has a flatter probability distribution, a variety of codebook embeddings might be sampled to represent the same data for its different details, which leads to an enriched codebook and an enhanced codebook usage.

\begin{figure}[t!]
    \centering
    \begin{subfigure}[b]{0.43\textwidth}
        \centering
        \includegraphics[width=0.95\textwidth]{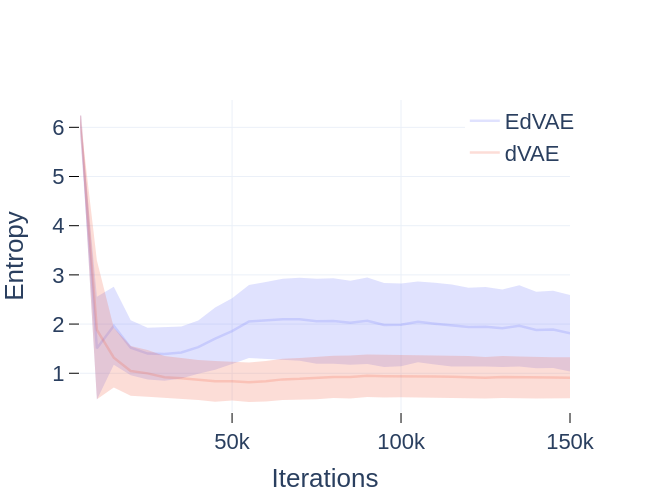}
        \caption{Average entropy of the probabilities during the training.}
        \label{fig:entropy_a}
    \end{subfigure}%
    \hfill
    \begin{subfigure}[b]{0.55\textwidth}
        \centering
        \includegraphics[width=0.95\textwidth]{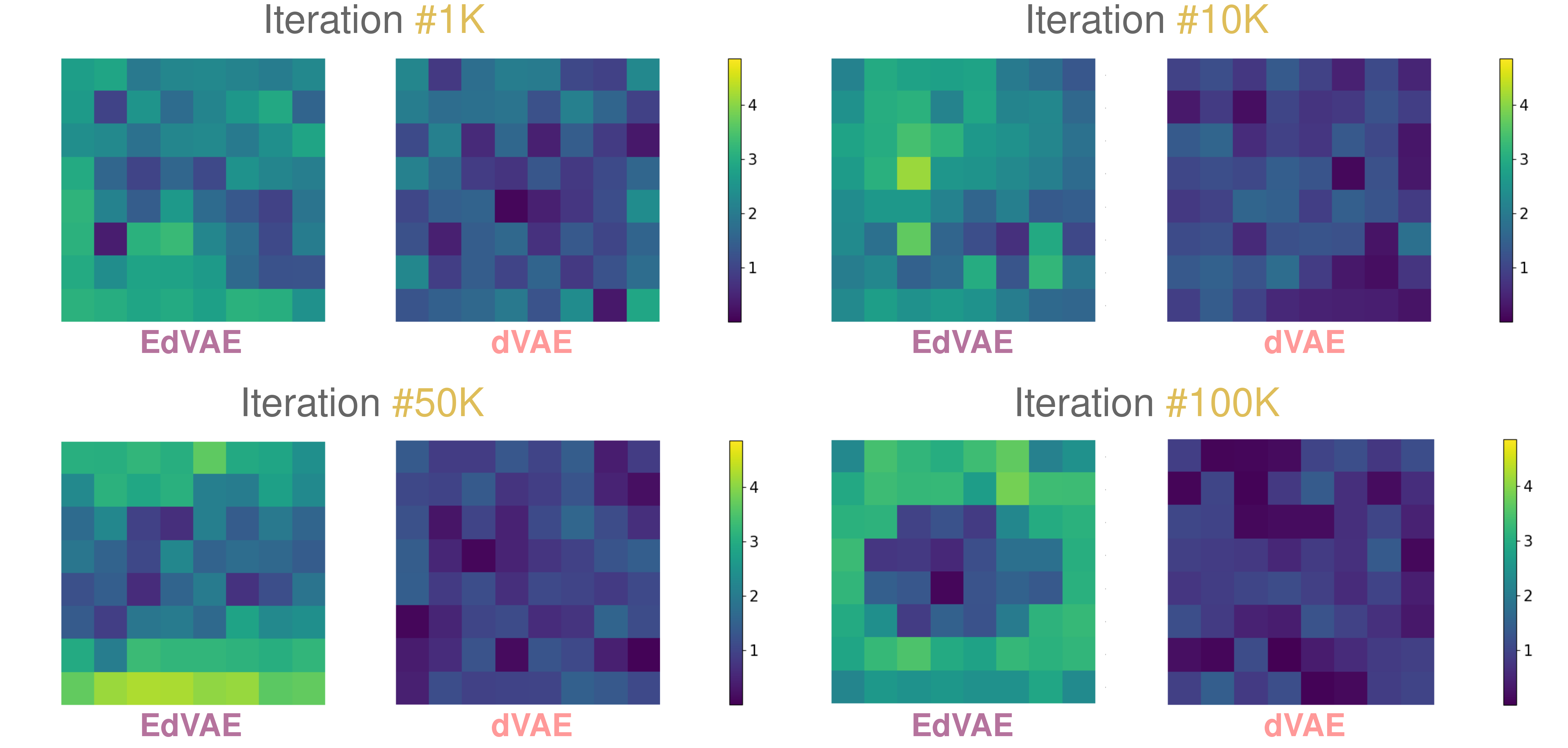}
        \caption{Entropy of the probabilities for each spatial position in the same sample during the training.}
        \label{fig:entropy_b}
    \end{subfigure}
    \caption{Entropy visualization of the probability distributions for CIFAR10.}
    \label{fig:entropy}
\end{figure}

Figure~\ref{fig:entropy_a} visualizes the average entropy of the probabilities during the training. We measure the entropy of each probability distribution over the codebook embeddings sample-wise, meaning that each sample consists of $N \times N$ number of entropy values calculated for each spatial position. We gather all $L \times N \times N$ entropy values at on the entire dataset where $L$ is the number of training samples, and plot the average entropy of the probabilities with mean and standard deviation. We find out that EdVAE's mean values of the entropy are higher than those of dVAE's, and this performance gain in terms of entropy is preserved during the training. Furthermore, we observe higher standard deviation for EdVAE in contrast to dVAE, which indicates that dVAE squeezes the probability masses into a smaller interval for all positions while the entropy values of EdVAE have a wider range aiding a relatively liberal codebook usage. 

Figure~\ref{fig:entropy_b} visualizes the entropy changes of the probabilities for each spatial position in the same sample during the training. All iterations have the same color bar for both models to observe the change effectively. Heat map visualization is useful to monitor the entropy of the probabilities within a single sample, and EdVAE obtains higher entropy values for most of the spatial positions compared to those of dVAE during the training. 

We note that while the prior works \citep{williams2020hierarchicalqa, takida2022sqvae} induce high entropy via regularizers, the feature introduced by our ELBO formulation inherently achieves the same effect as a result of our modeling assumptions that harness the power of the Dirichlet distribution.

While the spiky softmax distribution is the main problem, one might think to increase the stochasticity of sampling from the Categorical distribution with a higher temperature value to reduce the effects of overestimated probabilities. In order to test this assumption, we use higher temperature values with dVAE on CIFAR10 and CelebA datasets (see Appendix~\ref{appendix:temp_exp} for details). We show that the perplexity does not increase with a high temperature, and including randomness insensibly does not always affect the training in a good way. Our method incorporates stochasticity such that we learn how to perturb probabilities from the input. Learning a distribution over Categorical distributions from the input directly makes the model more reliable and resistant to hyperparameter change. 

\subsubsection{Uncertainty vs Perplexity}\label{subsec:uncertainty_effect}

\begin{figure}[t!]
    \centering
    \includegraphics[width=0.8\textwidth]{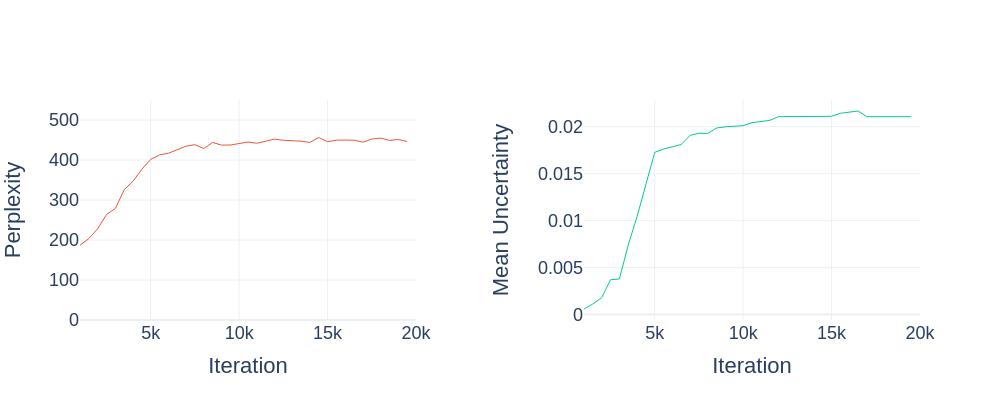}
    \caption{EdVAE training on CIFAR10: perplexity values increase during the training due to the increase in uncertainty values.}
    \label{fig:uncertainty}
\end{figure}

We anticipate that introducing uncertainty awareness will enhance codebook usage, addressing limitations in dVAE caused by the softmax operation. Our intuition is rooted in the definition of codebook collapse—new elements are introduced when existing ones fail to explain observations. In order to validate our intuition, we monitor the training of CIFAR10 and present an interval of the training until saturation in Figure~\ref{fig:uncertainty}.

We observe a correlation between the perplexity values and the uncertainty values during the training. The trend of perplexity values perfectly matches the trend of uncertainty values. This correlation emphasizes that our model dynamically adjusts codebook usage based on its uncertainty, preventing codebook collapse by utilizing embeddings effectively.

\subsubsection{Perplexity and reconstruction performance}\label{subsec:performance}

\begin{table}[t!]
\caption{Comparison of the models in terms of perplexity ($\uparrow$) using a codebook $\mathcal{M}^{512 \times 16}$.}
\label{tab:eval_perplexity}
\centering
\begin{tabular}{llll}
    \hline
    \multicolumn{1}{l}{\textbf{Method}} &\multicolumn{1}{l}{\textbf{CIFAR10}} &\multicolumn{1}{l}{\textbf{CelebA}} &\multicolumn{1}{l}{\textbf{LSUN Church}}\\
    \hline
    VQ-VAE-EMA \citep{oord2017neural} & 412.67 $\pm$ 2.05 & 405.33 $\pm$ 5.88 & 379.67 $\pm$ 3.09  \\
    GS-VQ-VAE \citep{Snderby2017ContinuousRT} & $ 208.33 \pm 6.03$ & $ 193.33 \pm 10.68$ & $ 189.67 \pm 5.02$ \\
    SQ-VAE \citep{takida2022sqvae} &  407.33 $\pm$ 7.32  & $\bf 409.33 \pm 2.05$ & 374.00 $\pm$ 2.16\\
    VQ-STE++ \citep{huh2023improvedvqste} & 414.33 $\pm$ 9.10 & 370.33 $\pm$ 4.11 & 375.67 $\pm$ 5.58\\
    dVAE \citep{ramesh2021zeroshot} & $ 190.33 \pm 13.02$ & $ 254.67 \pm 11.08$ & $ 363.33 \pm 4.07$ \\
    \hline
    EdVAE  & $\bf 420.33 \pm 4.49$ & $ 371.33 \pm 2.86$ & $\bf 385.67 \pm 5.63$  \\
    \hline
\end{tabular}
\end{table}

\begin{table}[t!]
\caption{Comparison of the models in terms of MSE ($\times 10^3$, $\downarrow$) using a codebook $\mathcal{M}^{512 \times 16}$.}
\label{tab:eval_mse}
\centering
\begin{tabular}{llll}
    \hline
    \multicolumn{1}{l}{\textbf{Method}} &\multicolumn{1}{l}{\textbf{CIFAR10}} &\multicolumn{1}{l}{\textbf{CelebA}} &\multicolumn{1}{l}{\textbf{LSUN Church}}\\
    \hline
    VQ-VAE-EMA \citep{oord2017neural} & 3.21 $\pm$ 0.05 & $ 1.07 \pm 0.06$ & 1.71 $\pm$ 0.05 \\
    GS-VQ-VAE \citep{Snderby2017ContinuousRT} &$ 3.63 \pm 0.01$ & $ 1.32 \pm 0.02$ & $ 1.84 \pm  0.06$\\
    SQ-VAE \citep{takida2022sqvae} & 4.01 $\pm$ 0.03 & 1.05 $\pm$ 0.02 & 1.79 $\pm$ 0.03\\
    VQ-STE++ \citep{huh2023improvedvqste} & 3.82 $\pm$ 0.1 & 1.11 $\pm$ 0.08 & 1.83 $\pm$ 0.03 \\
    dVAE \citep{ramesh2021zeroshot}  & $ 3.42 \pm  0.08 $ & $ 1.01 \pm 0.08$ & $ 1.60 \pm 0.01$\\
    \hline
    EdVAE  & $\bf 2.99 \pm 0.04$ & $\bf 0.89 \pm 0.01$ & $\bf 1.58 \pm 0.01$ \\
    \hline
\end{tabular}
\end{table}

\begin{figure}[t!]
   \centering
    \begin{subfigure}[b]{0.48\textwidth}
        \centering
        \includegraphics[width=\textwidth]{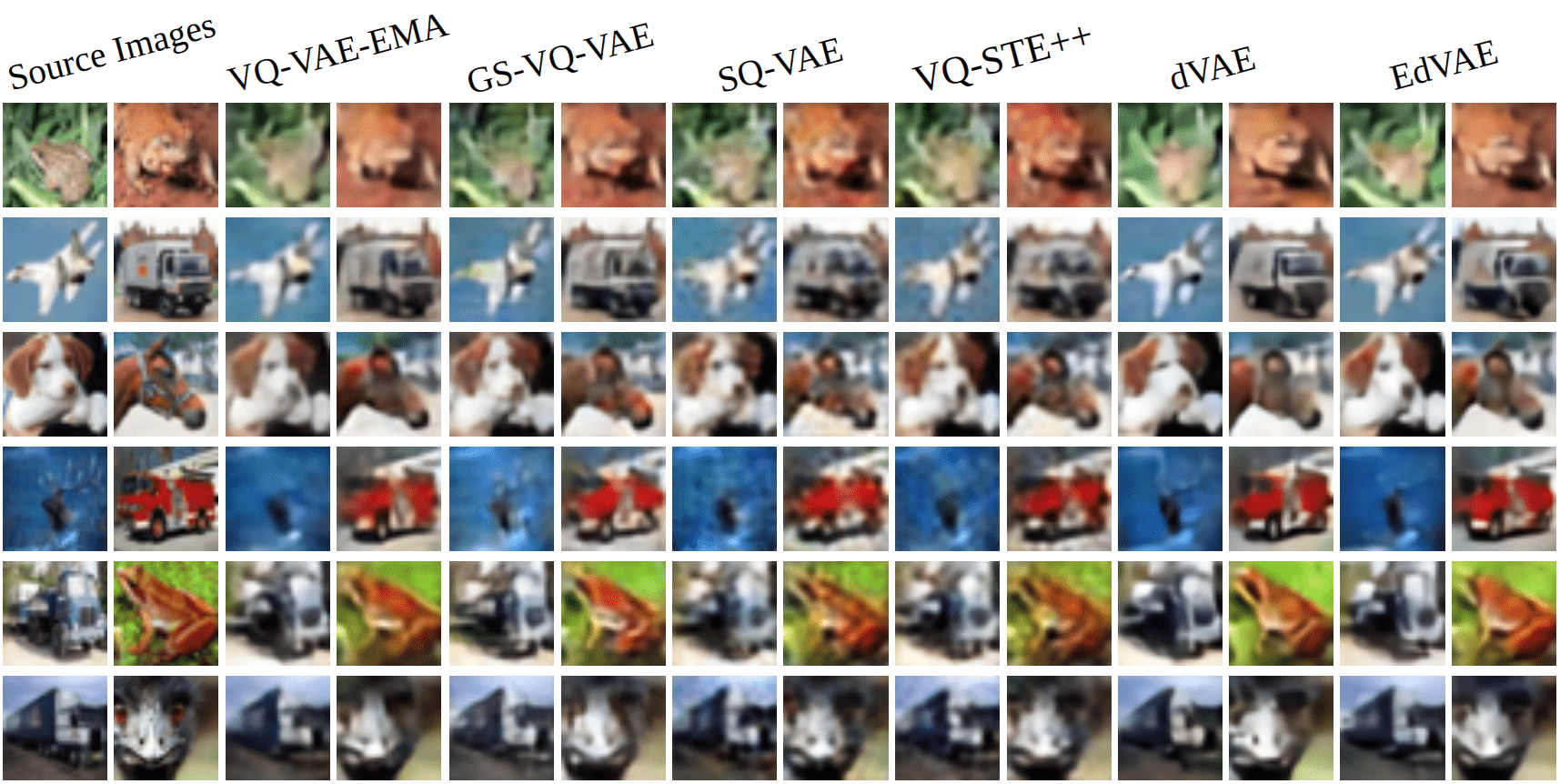}
        \caption{CIFAR10.}
        \label{fig:cifar_comparison}
    \end{subfigure}
    \hfill
    \begin{subfigure}[b]{0.48\textwidth}
        \centering
        \includegraphics[width=\textwidth]{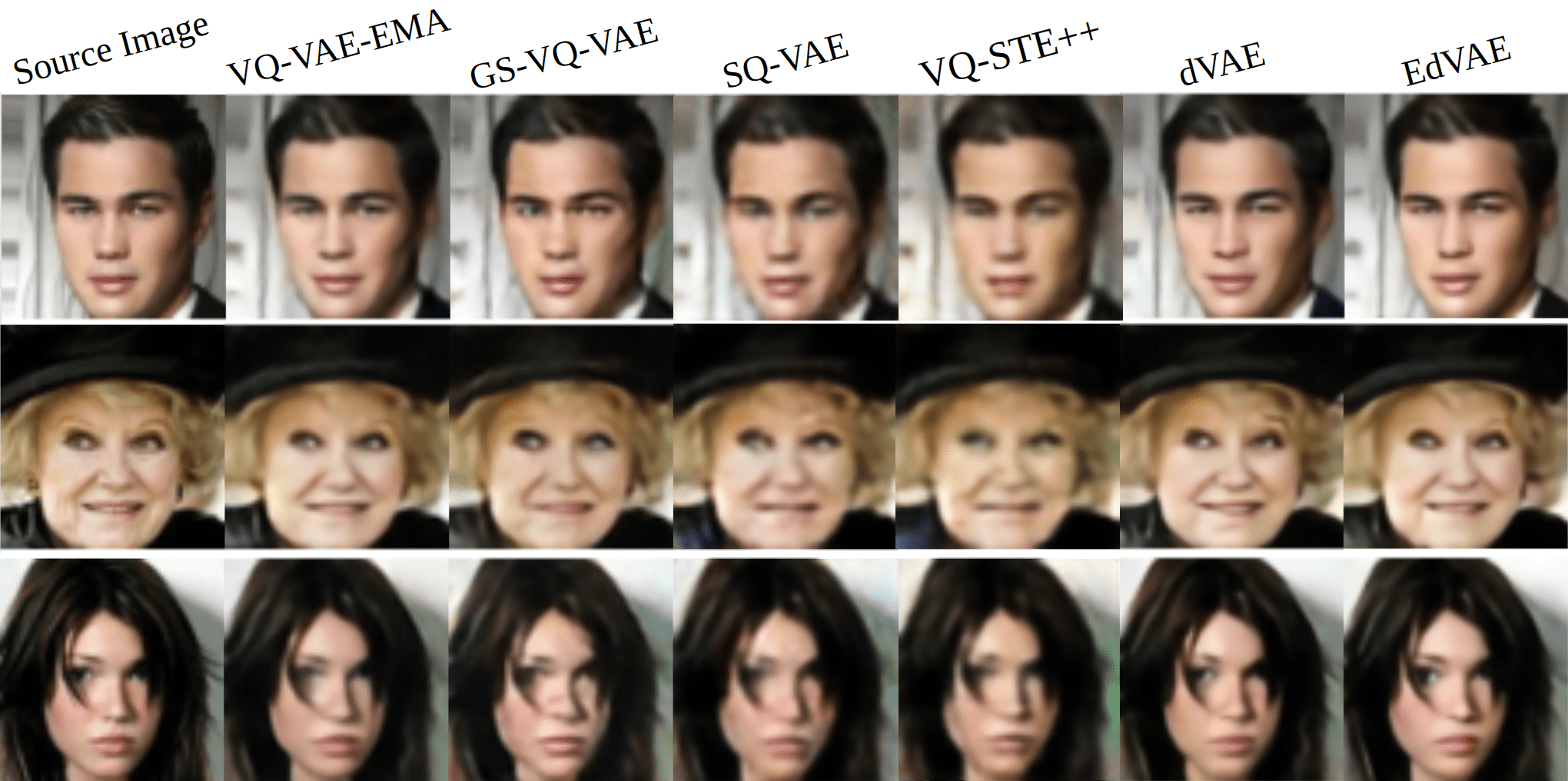}
        \caption{CelebA.}
        \label{fig:celeba_comparison}
    \end{subfigure}  
   \caption{Reconstructions from (a) CIFAR10, (b) CelebA.}
\end{figure}

\begin{figure}[t!]
   \centering
    \includegraphics[width=0.6\textwidth]{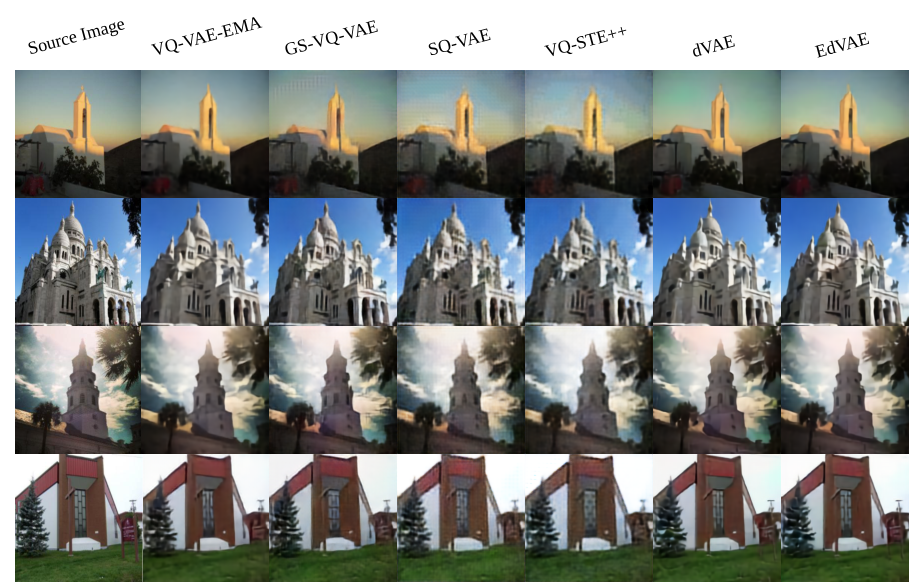}
    \caption{Reconstructions from LSUN Church.}
    \label{fig:lsun_comparison}
\end{figure}

In order to evaluate the codebook usage of all models, we leverage on the perplexity metric whose upper bound is equal to the number of the codebook embeddings. Therefore, we directly compare the number of used embeddings for all models. Perplexity results of all models are presented in Table~\ref{tab:eval_perplexity}. Additionally, since the reconstruction performance should not be lost while increasing the diversity of the codebook embeddings used in the latent representation, we also evaluate all models in terms of MSE. Numerical results are presented in Table~\ref{tab:eval_mse}. While we obtain the lowest MSE values for all datasets, we outperform the other methods for CIFAR10 and LSUN Church datasets in terms of perplexity. Whereas SQ-VAE achieves the highest perplexity result for CelebA dataset, EdVAE performs close to SQ-VAE's perplexity, while obtaining a better reconstruction performance than SQ-VAE and the other methods. Moreover, EdVAE outperforms dVAE substantially in perplexity. It is important to note that EdVAE not only mitigates the codebook collapse problem of dVAE, but also outperforms the VQ-VAE based methods. 

We evaluate our model and the other models visually for a qualitative evaluation assessment. Figure~\ref{fig:cifar_comparison}, Figure~\ref{fig:celeba_comparison}, and Figure~\ref{fig:lsun_comparison} compare all of the models' reconstruction performance on CIFAR10, CelebA, and LSUN Church datasets, respectively. We observe that while our model reconstructs most of the finer details such as the shape of eye or gaze direction better than the other models in CelebA, we also include some examples where some of the finer details such as the shape of the nose and the mouth are depicted by one of the opponent models better than our model. For LSUN Church examples, we observe that our model's reconstructions depict the colors and the shapes in the source images better than the other models.

\subsubsection{Effects of codebook design}\label{subsec:codebook_design}

\begin{figure}[t!]
    \centering
    \includegraphics[width=0.9\textwidth]{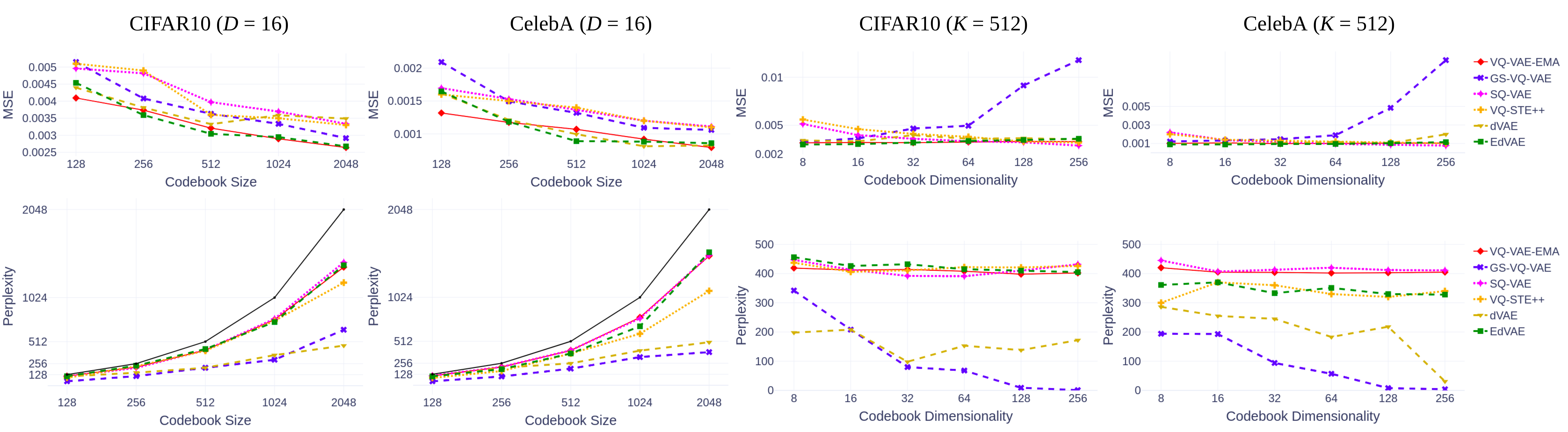}
    \caption{Impact of codebook design on perplexity and MSE, using CIFAR10 and CelebA datasets. The black ``codebook size" line indicates the upper bound for the perplexity.}
    \label{fig:codebook_design}
\end{figure}

We emphasize the critical role of codebook design using CIFAR10 and CelebA datasets in Figure~\ref{fig:codebook_design}. It is important to have a model that uses most of the codebook embeddings even with a larger codebook. Therefore, we evaluate EdVAE's and other methods' performance using various codebook sizes and dimensionalities. In order to observe the effects of size and dimensionality separately, we fix the dimensionality to 16 while we use different codebook sizes ranging from 128 to 2048. Then, we fix the size to 512 while we use different codebook dimensionalities ranging from 8 to 256.

Figure~\ref{fig:codebook_design} shows that EdVAE outperforms the other methods in most of the setting where we use different codebook sizes. EdVAE's perplexity is not affected negatively when the codebook size increases, and it obtains the lowest MSE values in most of the settings for both of the datasets. Therefore, EdVAE is suitable to work with a larger codebook. When the codebook dimensionality changes, we observe that EdVAE outperforms the other methods or obtains compatible results. While the other methods seem sensitive to the codebook design, EdVAE does not need a carefully designed codebook which eases the process of model construction.

\subsubsection{Approximated prior}\label{subsec:generation}

\begin{table}[t!]
\caption{Comparison of the models in terms of FID ($\downarrow$).}
\label{tab:fid_table}
\centering
\begin{tabular}{cccc}
    \hline
    \multicolumn{1}{c}{\textbf{Method}} &\multicolumn{1}{c}{\textbf{CIFAR10}} &\multicolumn{1}{c}{\textbf{CelebA}} &\multicolumn{1}{c}{\textbf{LSUN Church}}\\
    \hline
    VQ-VAE-EMA \citep{oord2017neural} & $57.04 \pm 2.32$ & $ 34.30 \pm 2.41 $ & $71.22 \pm 2.72$ \\
    GS-VQ-VAE \citep{Snderby2017ContinuousRT} & $ 56.35 \pm 2.17$ & $ 33.12 \pm 1.30 $ & $ 72.52 \pm 3.25 $\\
    SQ-VAE \citep{takida2022sqvae} & $54.17 \pm 2.85$  &  $ 33.03 \pm 1.04 $  &  $ \bf 63.41 \pm 2.36$ \\
    VQ-STE++ \citep{huh2023improvedvqste} & $ 55.53 \pm 1.97$ & $ 32.98 \pm 2.27 $ & $ 71.03 \pm 1.95 $  \\
    dVAE \citep{ramesh2021zeroshot} & $ 58.85 \pm 0.93 $ & $ 37.29 \pm 3.14 $ & $ 71.32 \pm 0.71 $\\
    \hline
    EdVAE  & $ \bf 51.82 \pm 1.58 $ & $ \bf 32.51 \pm 1.13 $ & $69.63 \pm 1.29 $\\
    \hline
\end{tabular}
\end{table}

\begin{table}[t!]
\caption{Comparison of the models in terms of Precision \& Recall ($\uparrow$).}
\label{tab:pr_table}
\centering
\begin{tabular}{cccc}
    \hline
    \multicolumn{1}{c}{\textbf{Method}} &\multicolumn{1}{c}{\textbf{CIFAR10}} &\multicolumn{1}{c}{\textbf{CelebA}} &\multicolumn{1}{c}{\textbf{LSUN Church}}\\
    \hline
    VQ-VAE-EMA \citep{oord2017neural} & $ 0.47, 0.32 $ & $ 0.44, 0.35 $ & $ 0.19, 0.15 $ \\
    GS-VQ-VAE \citep{Snderby2017ContinuousRT} & $ 0.45, 0.34 $ & $ 0.39, 0.32 $ & $ 0.20, 0.14 $\\
    SQ-VAE \citep{takida2022sqvae} & $0.52, 0.30 $  &  $ \bf 0.52, 0.37 $  & $ 0.23, 0.17 $ \\
    VQ-STE++ \citep{huh2023improvedvqste} & $ 0.51, 0.32 $ & $ 0.47, 0.36 $ & $ 0.24, 0.15 $  \\
    dVAE \citep{ramesh2021zeroshot} & $ 0.43, 0.34 $ & $ 0.41, 0.30 $ & $ 0.20, 0.15 $  \\
    \hline
    EdVAE  & $ \bf 0.54, 0.35 $ & $ 0.48, 0.35 $ & $ \bf 0.28, 0.18 $  \\
    \hline
\end{tabular}
\end{table}


\begin{figure}[t!]
    \centering
    \begin{subfigure}[b]{0.30\textwidth}
        \centering
        \includegraphics[width=0.98\textwidth]{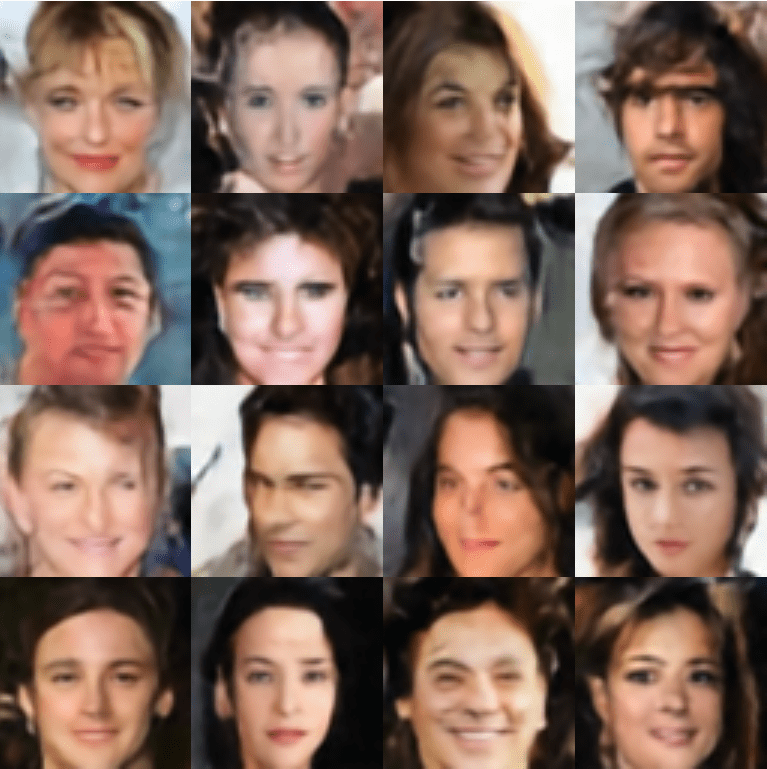}
        \caption{VQ-VAE-EMA}
    \end{subfigure}
    \hfill
    \begin{subfigure}[b]{0.30\textwidth}
        \centering
        \includegraphics[width=0.98\textwidth]{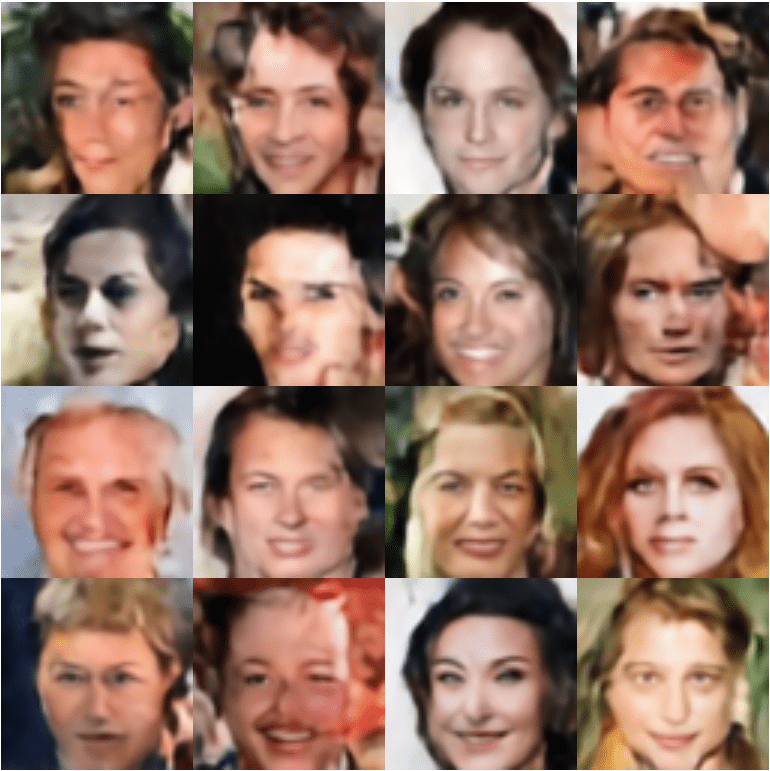}
        \caption{GS-VQ-VAE}
    \end{subfigure}
    \hfill
    \begin{subfigure}[b]{0.30\textwidth}
        \centering
        \includegraphics[width=0.98\textwidth]{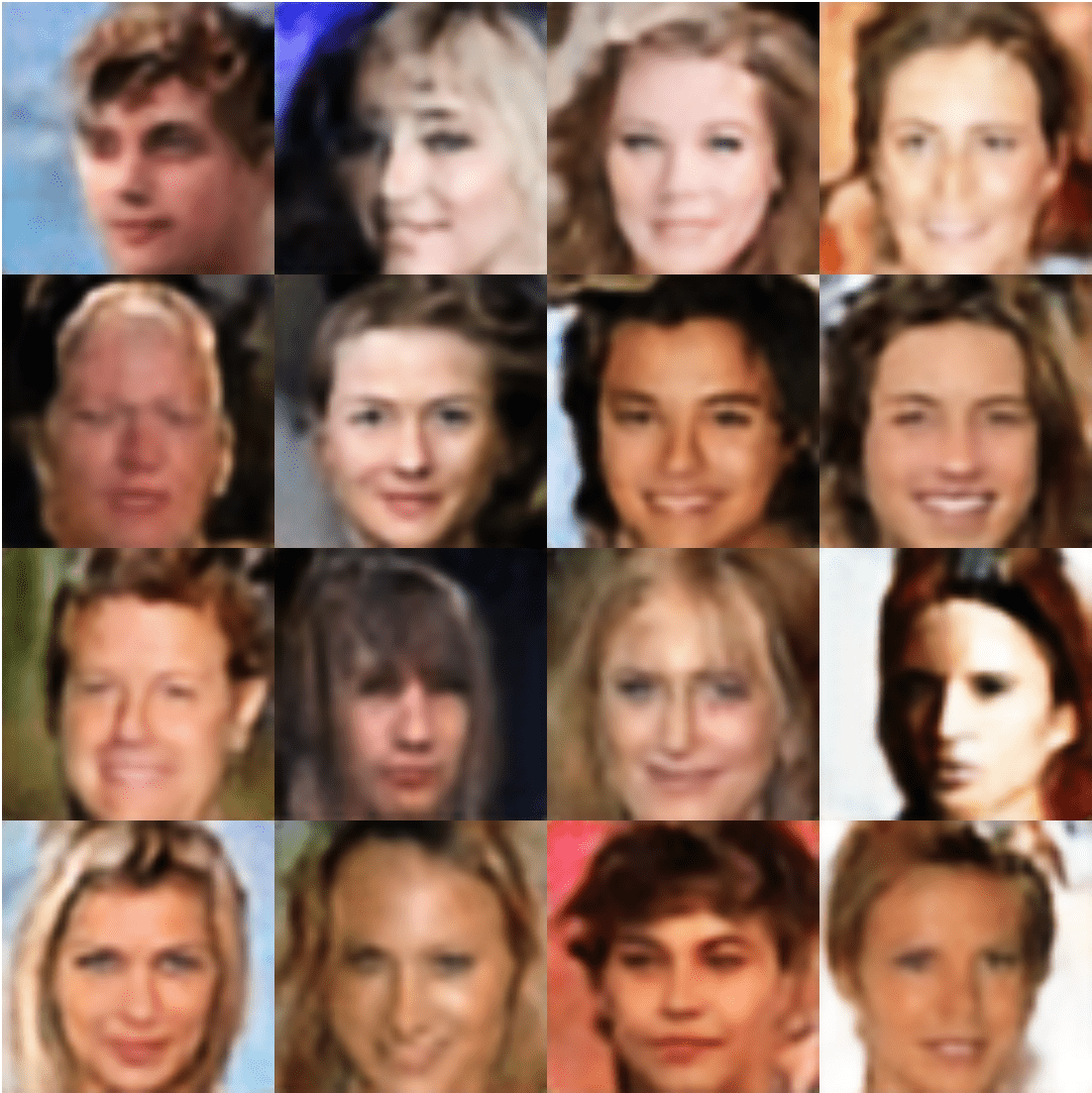}
        \caption{SQ-VAE}
    \end{subfigure}
    \hfill
    \begin{subfigure}[b]{0.30\textwidth}
        \centering
        \includegraphics[width=0.98\textwidth]{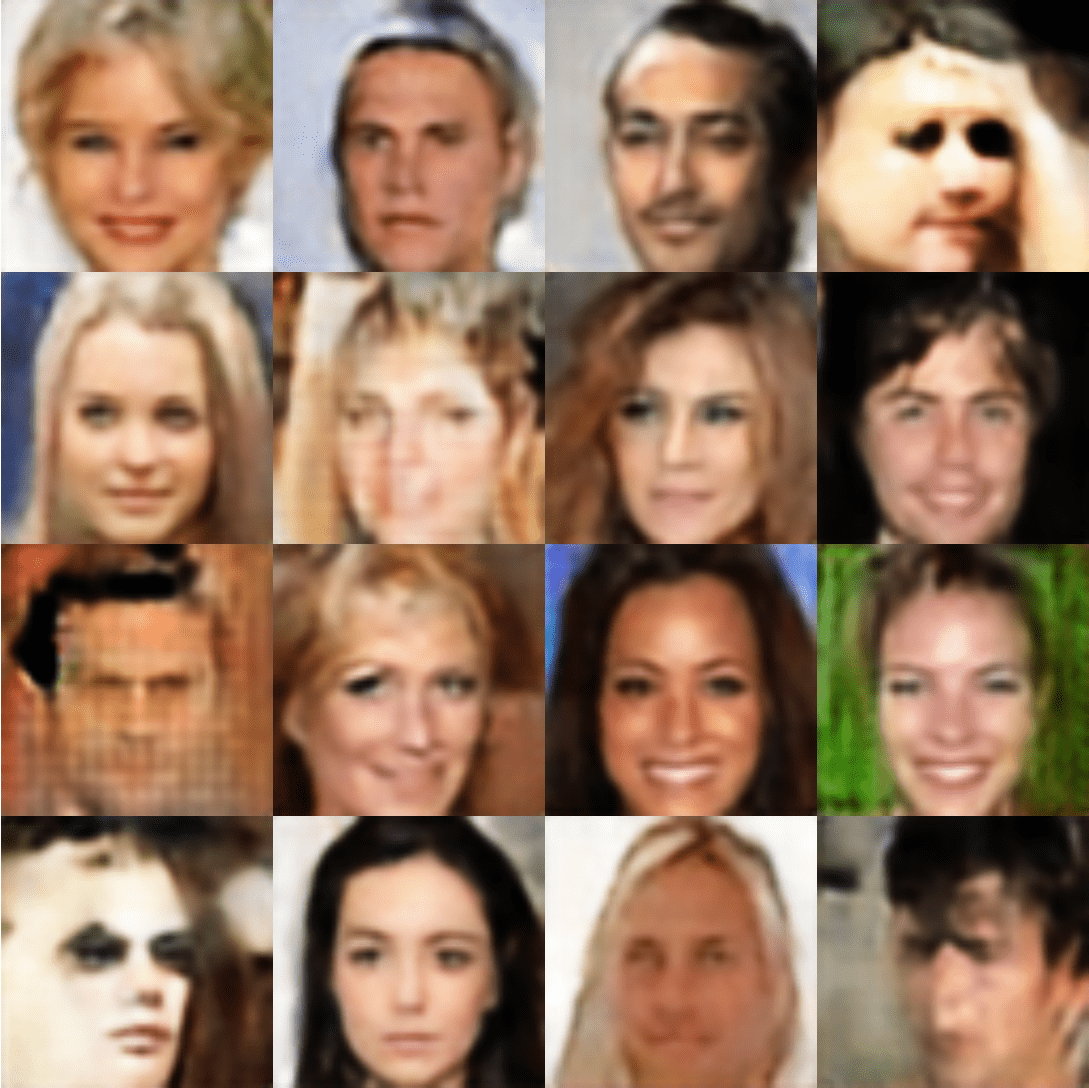}
        \caption{VQ-STE++}
    \end{subfigure}
    \hfill
    \begin{subfigure}[b]{0.30\textwidth}
        \centering
        \includegraphics[width=0.98\textwidth]{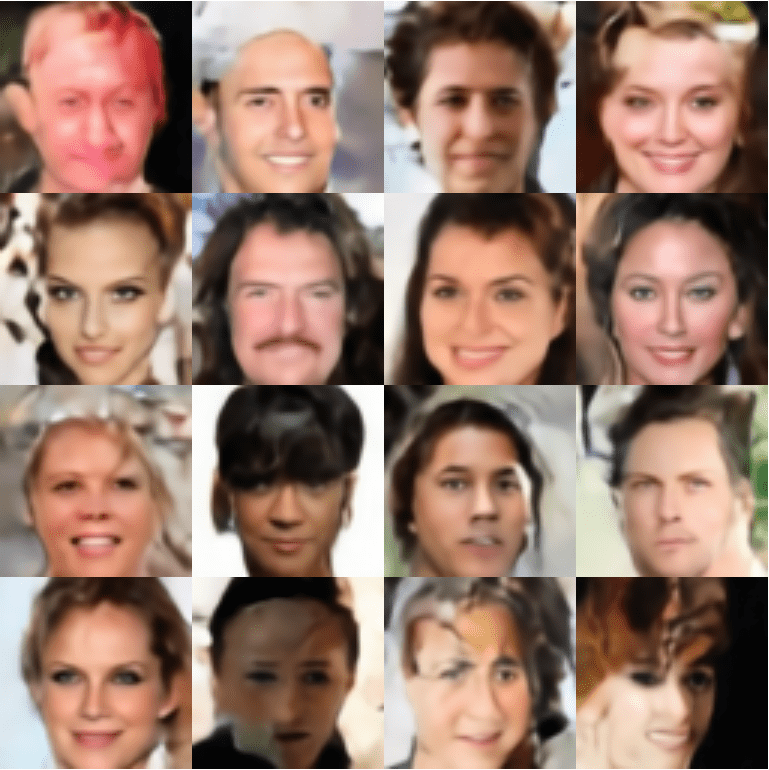}
        \caption{dVAE}
    \end{subfigure}
    \hfill
    \begin{subfigure}[b]{0.30\textwidth}
        \centering
        \includegraphics[width=0.98\textwidth]{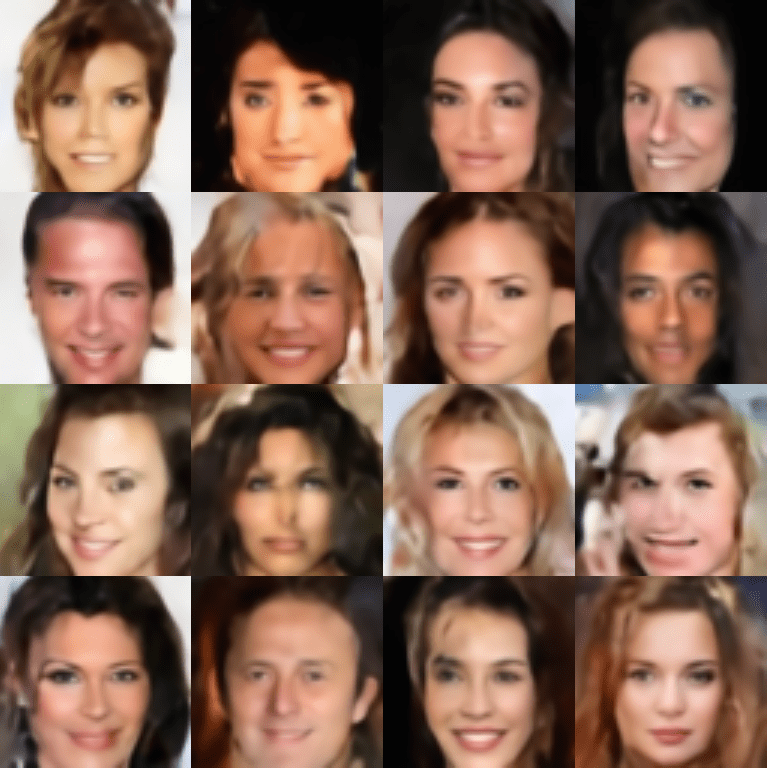}
        \caption{EdVAE}
    \end{subfigure}
    \caption{Generated samples from CelebA dataset.}
    \label{fig:celeb_generation_all}
\end{figure}

\begin{figure}[t!]
    \centering
    \begin{subfigure}[b]{0.30\textwidth}
        \centering
        \includegraphics[width=0.98\textwidth]{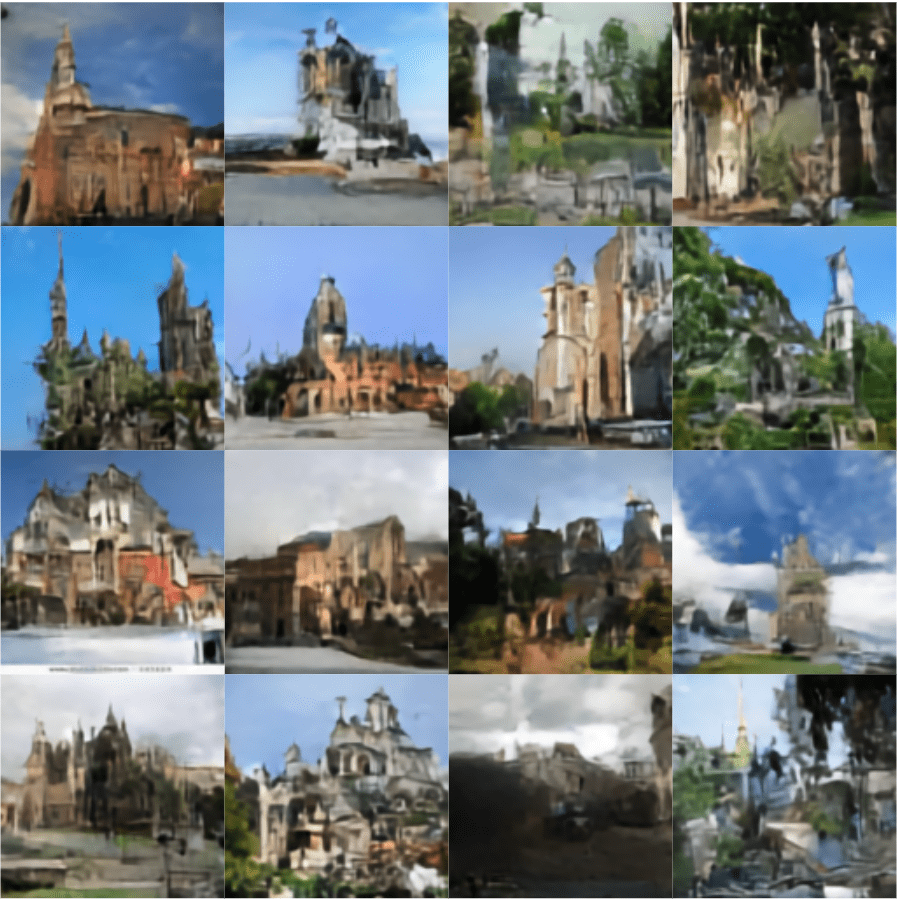}
        \caption{VQ-VAE-EMA}
    \end{subfigure}
    \hfill
    \begin{subfigure}[b]{0.30\textwidth}
        \centering
        \includegraphics[width=0.98\textwidth]{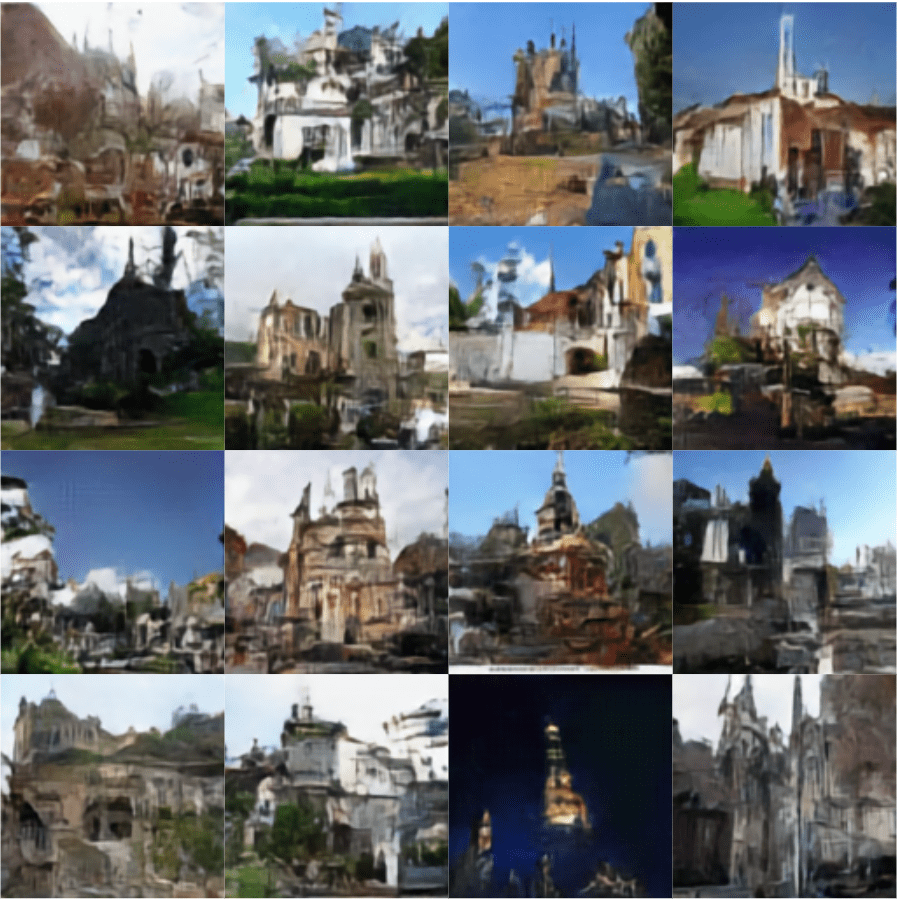}
        \caption{GS-VQ-VAE}
    \end{subfigure}
    \hfill
    \begin{subfigure}[b]{0.30\textwidth}
        \centering
        \includegraphics[width=0.98\textwidth]{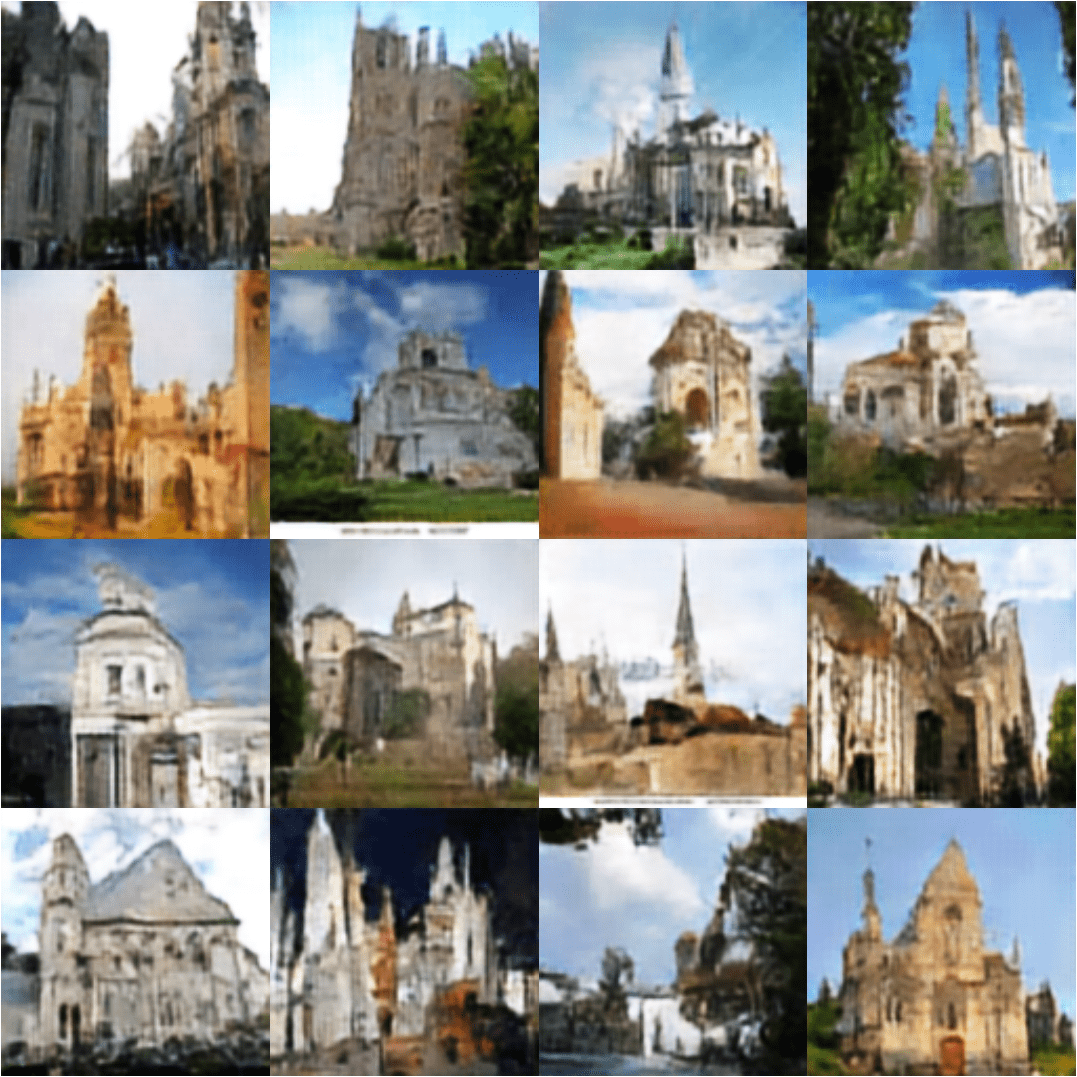}
        \caption{SQ-VAE}
    \end{subfigure}
    \hfill
    \begin{subfigure}[b]{0.30\textwidth}
        \centering
        \includegraphics[width=0.98\textwidth]{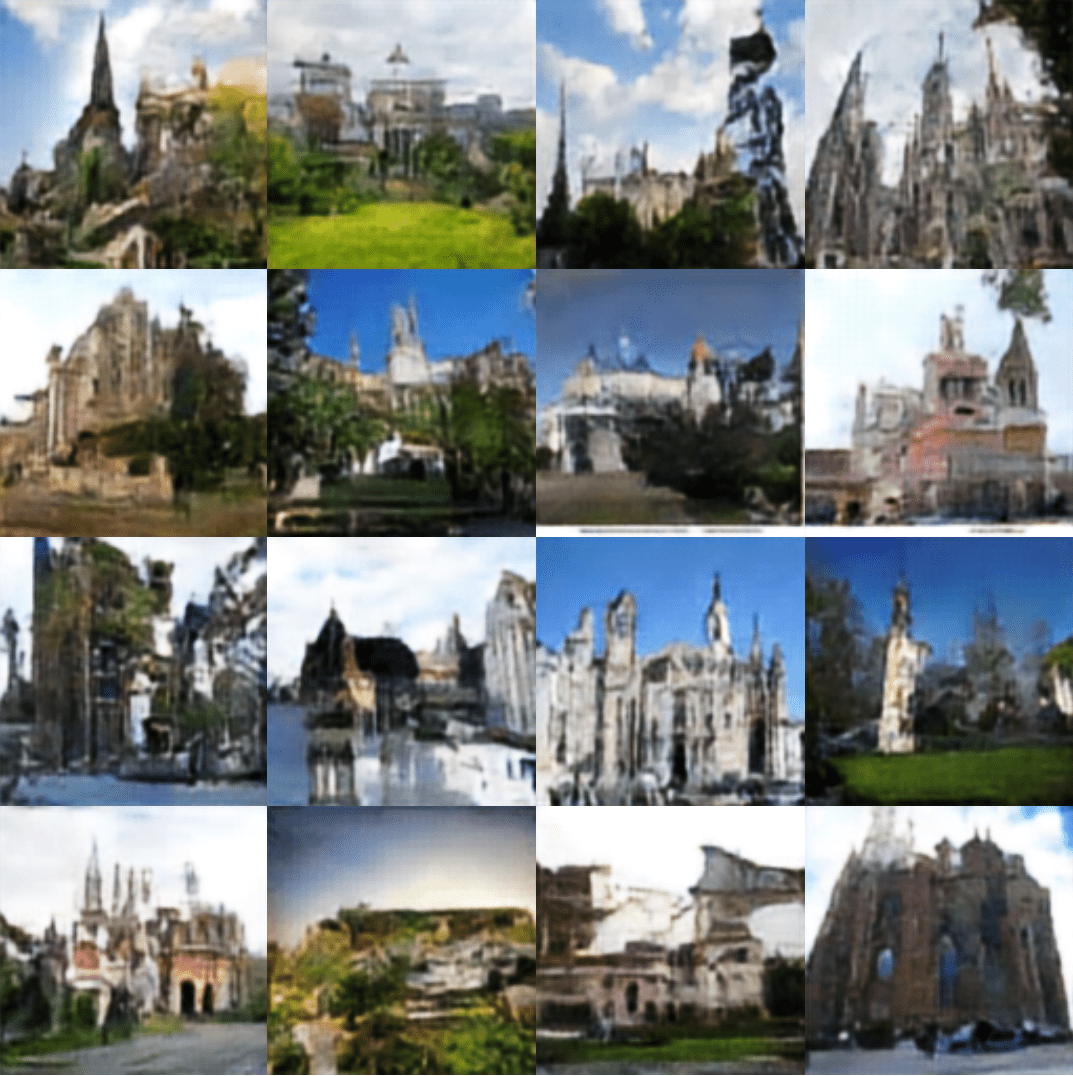}
        \caption{VQ-STE++}
    \end{subfigure}
    \hfill
    \begin{subfigure}[b]{0.30\textwidth}
        \centering
        \includegraphics[width=0.98\textwidth]{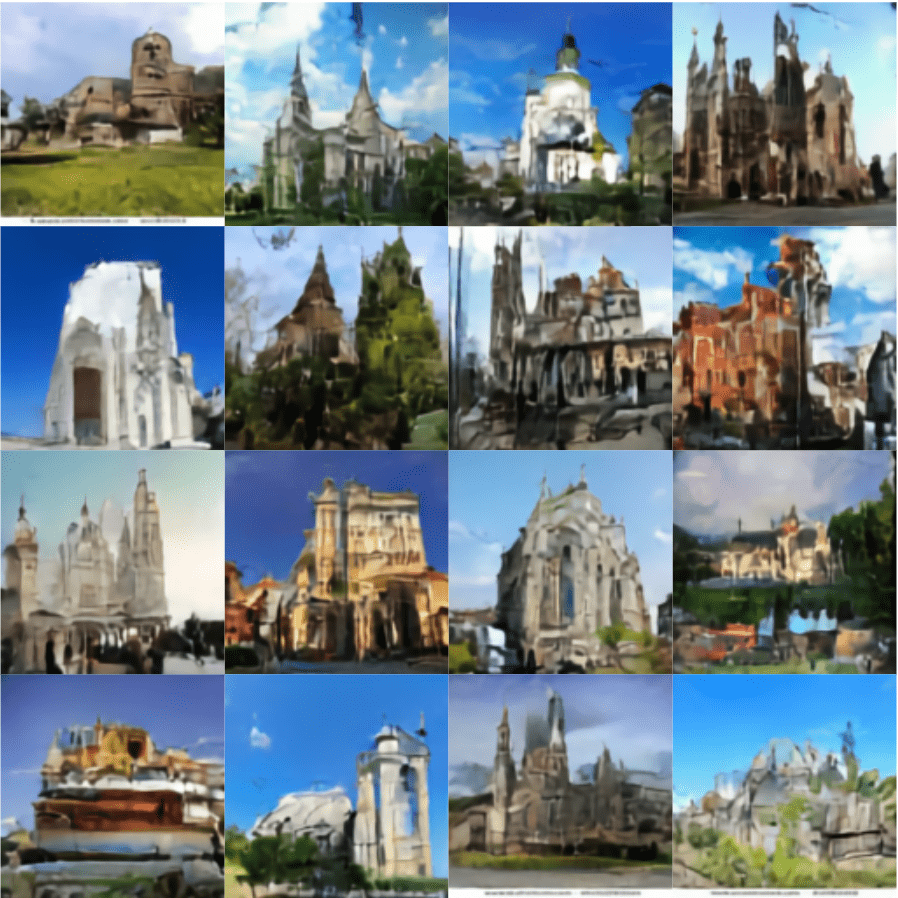}
        \caption{dVAE}
    \end{subfigure}
    \hfill
    \begin{subfigure}[b]{0.30\textwidth}
        \centering
        \includegraphics[width=0.98\textwidth]{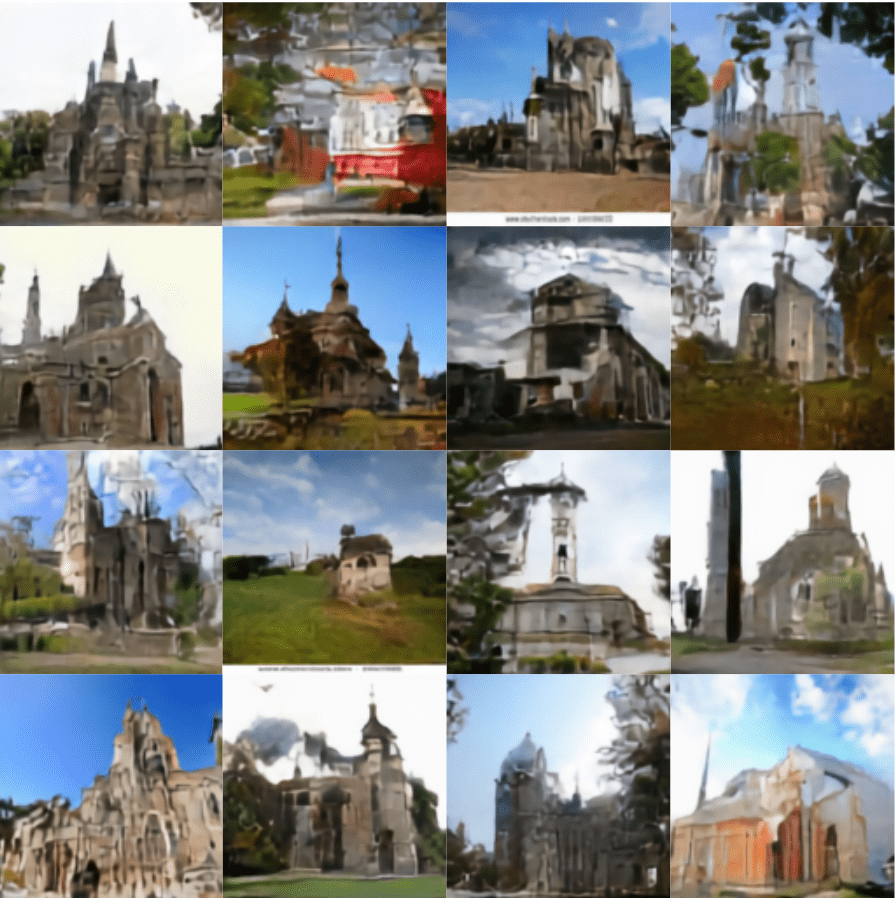}
        \caption{EdVAE}
    \end{subfigure}
    \caption{Generated samples from LSUN Church dataset.}
    \label{fig:lsun_generation_all}
\end{figure}

\begin{figure}[t!]
    \centering
    \begin{subfigure}[b]{0.15\textwidth}
        \centering
        \includegraphics[width=0.95\textwidth]{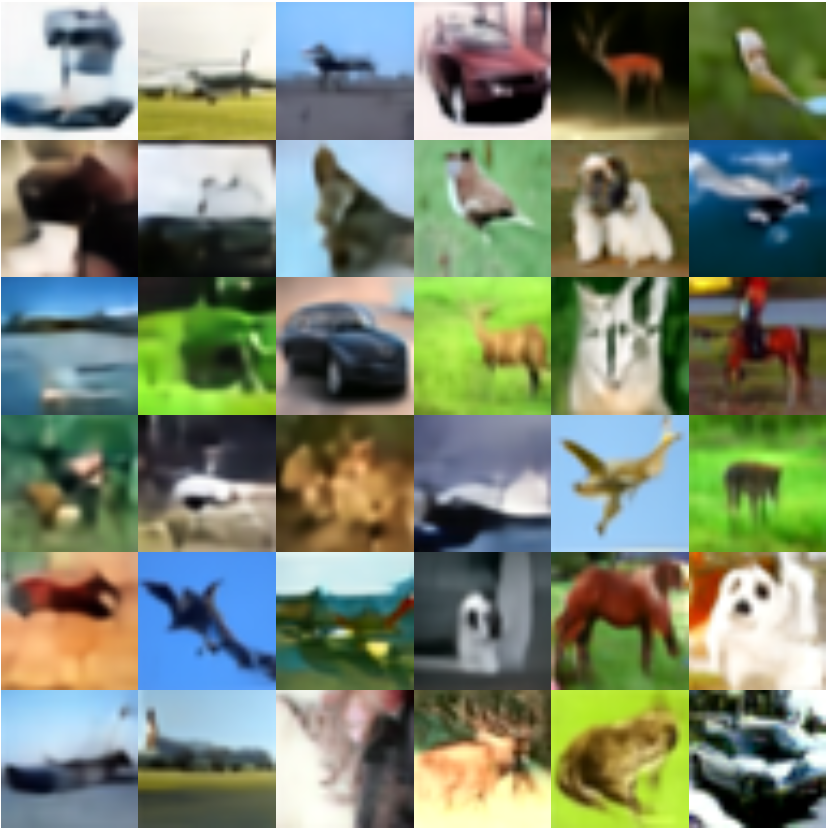}
        \caption{CIFAR10.}
    \end{subfigure}%
    \hfill
    \begin{subfigure}[b]{0.41\textwidth}
        \centering
        \includegraphics[width=0.95\textwidth]{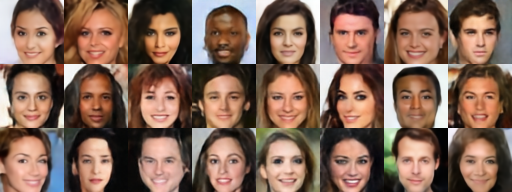}
        \caption{CelebA.}
    \end{subfigure}%
    \hfill
    \begin{subfigure}[b]{0.41\textwidth}
        \centering
        \includegraphics[width=0.95\textwidth]{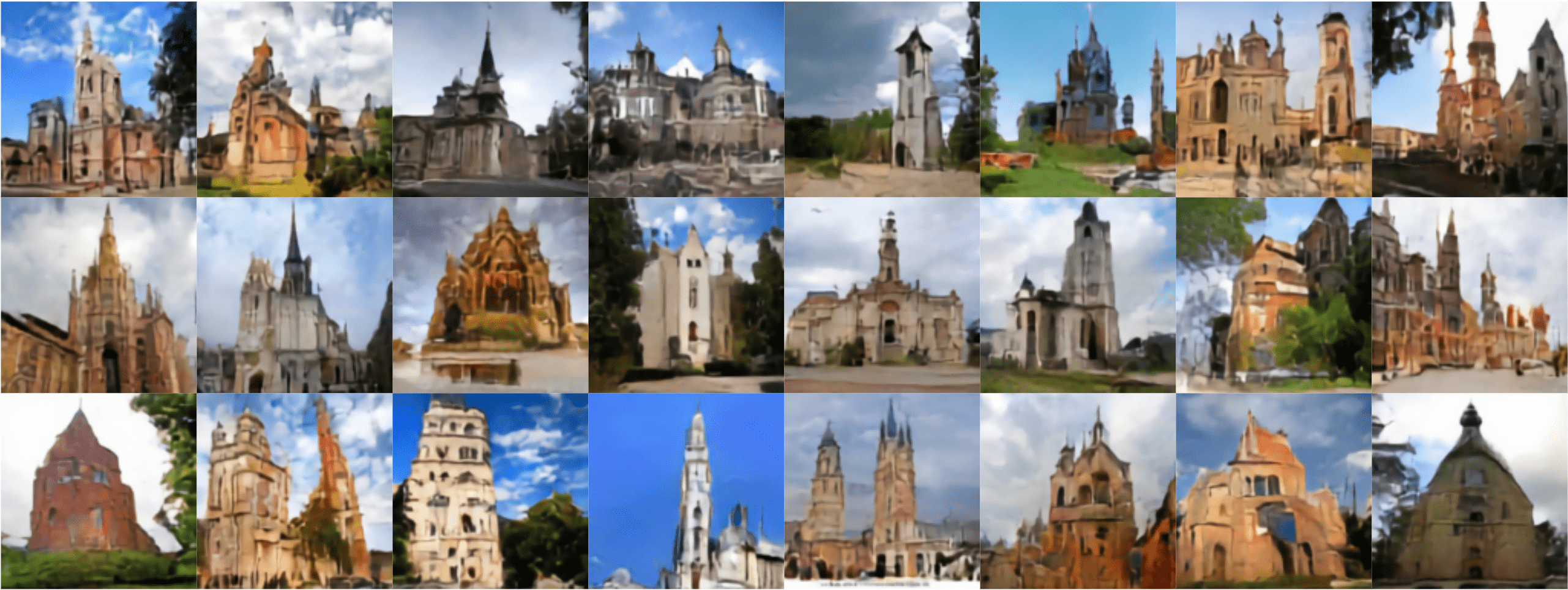}
        \caption{LSUN Church.}
    \end{subfigure}
    \caption{Generated samples using the learned prior over the discrete indices produced by EdVAE.}
    \label{fig:generation_results}
\end{figure}

Image generation is a downstream task over which we evaluate the performance of the discrete latent spaces learned by the discrete VAEs. As the prior used during the training of $\mathcal{E}_\theta$, $\mathcal{D}_\phi$, and $\mathcal{M}$ is a uniform distribution, i.e. an uninformative prior, it should be updated to accurately reflect the true distribution over the discrete latents. Therefore, we fit an autoregressive distribution over the discrete latents of the training samples, for which we follow PixelSNAIL \citep{chen2018pixelsnail}. PixelSNAIL is also used in VQ-VAE-2 instead of PixelCNN \citep{oord2016pixel} that is used in VQ-VAE and SQ-VAE. 

We report FID and Precision-Recall in Table~\ref{tab:fid_table} and Table~\ref{tab:pr_table} to evaluate the quality of the generated images that are created autoregressively after a training over the discrete codebook indices attained from a given model, respectively. The FID results imply that EdVAE performs better than the other models in all datasets except LSUN Church, while it outperforms other models in terms of Precision-Recall in all datasets except CelebA. These results elucidate the point that the discrete latent spaces learned by EdVAE helps to attain a reinforced representation capacity for the latent space, which is a desirable goal for the image generation task. Figure~\ref{fig:celeb_generation_all} and Figure~\ref{fig:lsun_generation_all} present generated samples using the discrete latents of all models for CelebA and LSUN Church datasets, respectively. While all of the models manage to generate realistic images, structural consistency and semantic diversity in EdVAE's results are noticeable and in line with the numerical results. Figure~\ref{fig:generation_results} present additional EdVAE samples to advocate EdVAE's performance.

\section{Conclusion}\label{sec:conclusion}

The proposed EdVAE extends dVAE with a hierarchical Bayesian modeling to mitigate the codebook collapse problem of the latter. We demonstrate the essence of the problem, that is the confirmation bias caused by the spiky softmax distribution, and reformulate the optimization with an evidential view in order to acquire a hierarchy between the probability distributions aiding uncertainty-aware codebook usage. We show that our method outperforms the former methods in most of the settings in terms of reconstruction and codebook usage metrics.

Although the proposed method improves the codebook usage for dVAE family, there is still room for improvements both numerically and experimentally. We evaluate our method over small to medium size datasets compared to datasets like ImageNet \citep{deng2009imagenet}. As our work is the first to state the codebook collapse problem of dVAE and to propose a solution to it, it sets a baseline that can be extended to obtain improved outcomes in different models and datasets. We mainly provide evidence for the utility of our method for relatively structured datasets like CelebA or LSUN Church, and exploring the best parameterization for more diverse datasets like ImageNet would be a possible direction of our future work.

\section*{Acknowledgment}
In this work, Gulcin Baykal was supported by TÜB\.{I}TAK 2214-A International Research Fellowship Programme for PhD Students and Google DeepMind Scholarship Program at ITU. Computing resources used during this research were provided by the National Center for High Performance Computing of Turkey (UHeM) [grant number 1007422020].

\appendix
\section*{Appendix}
\renewcommand{\thesection}{\Alph{section}}

\section{Kullback-Leibler Divergence Between Two Dirichlet Distribution}\label{appendix:kl_derivation}

For two Dirichlet distributions $\mathcal{D}\textit{ir}(\pi | \alpha_\theta(x))$ and $\mathcal{D}\textit{ir}(\pi | 1,\ldots, 1)$ over $K$-dimensional probability $\pi$, the following equality holds
\begin{align}
    &\qquad \mathcal{D}_{\text{KL}}(\mathcal{D}\textit{ir}(\pi | \alpha_\theta(x)) || \mathcal{D}\textit{ir}(\pi | 1,\ldots, 1)) \nonumber \\
    &= \log \left( \frac{\Gamma\big(\sum_k \alpha_\theta^k(x)\big)}{\Gamma(K)\prod_k \Gamma(\alpha_\theta^k(x))}\right) + \sum_{k=1}^K (\alpha_\theta^k(x) - 1)\Big(\psi(\alpha_\theta^k(x)) - \psi\big(\textstyle\sum_k \alpha_\theta^k(x)\big)\Big). \nonumber 
    \label{eq:kl}
\end{align}
where $\Gamma(.)$ and $\psi(\alpha) := \frac{\mathrm d}{\mathrm d \alpha} \text{log} \Gamma(\alpha)$ are the \textit{gamma} and \textit{digamma} functions, respectively.

\section{Derivation Details}\label{appendix:derivations}

As we define in Section~\ref{sec:method}, our forward model is:
\begin{align}
    p(\pi) &= \mathcal{D}\textit{ir}(\pi | 1,\ldots, 1),\\
    Pr(z|\pi) &=\mathcal{C}\textit{at}(z|\pi), \\
    p(x|\mathcal{M},z=k) &=\mathcal{N}(x|\mathcal{D}_\phi(\mathcal{M}, z),\sigma^2 I).
\end{align}
and the approximate posterior is:
\begin{align}
    q(\pi, z|x) = \mathcal{C}\textit{at}(z|\pi) \mathcal{D}\textit{ir}(\pi | \alpha_\theta^1(x), \ldots, \alpha_\theta^K(x)).
\end{align}
We derive the ELBO to be maximized during the training as:
\begin{align}
    \log p(x|\mathcal{M},\theta,\phi) &= \log \mathbb{E}_{q(\pi, z|x )} \left [\frac{p(x|\mathcal{M},z) Pr(z|\pi) p(\pi)}{q(\pi, z |x )}\right]\\
    &= \log \mathbb{E}_{q(\pi, z |x )} \left [\frac{p(x|\mathcal{M},z) \cancel{Pr(z|\pi)} p(\pi)}{\cancel{Pr(z|\pi)} q(\pi|x)}\right]\\
    &\geq \mathbb{E}_{Pr(z |\pi)} \left [ \mathbb{E}_{q(\pi|x)} [ \log p(x|\mathcal{M},z)]  - \mathbb{E}_{q(\pi|x)}[\log q(\pi|x)/p(\pi)] \right ]\\
    &=\mathbb{E}_{Pr(z |\pi)} \left [ \mathbb{E}_{q(\pi|x)} [ \log p(x|\mathcal{M},z)] \right ]  - \mathcal{D}_{\text{KL}}(q(\pi|x) || p(\pi)). \label{eq:elbo_derivation}
\end{align}
The codebook can be viewed as $\mathcal{M} = [m_1, \ldots, m_K]$ where $m_k$s are the codebook embeddings. The input of the decoder $z_q(x) = \mathcal{M}_k = z * \mathcal{M}$ consists of the codebook embeddings $m_k$s as shown in Figure~\ref{fig:method}. $\mathcal{M}_k$ is retrieved using the indices $z$s sampled as $z \sim \mathcal{C}\textit{at}(z|\pi)$ via * operator which performs tensor-matrix multiplication. Therefore, we can use $\mathcal{D}_{\phi}(\mathcal{M}_k)$ instead of $\mathcal{D}_{\phi}(\mathcal{M}, z)$ for the remaining of the derivations.

The first term in Equation~\ref{eq:elbo_derivation} can be further derived as: 
\begin{align}
    \mathbb{E}&_{Pr(z |\pi)} \left [ \mathbb{E}_{q(\pi|x)} [ \log p(x|\mathcal{M},z)] \right ] = -\frac{1}{2}\log \sigma^2 - \frac{1}{2\sigma^2}   \sum_k q(\pi_k|x) (x-\mathcal{D}_{\phi}(\mathcal{M}_k))^2 \tag{\stepcounter{equation}\theequation}\\
    &\propto \mathbb{E}_{q(\pi|x)} \left [\sum_k \pi_k (x-\mathcal{D}_{\phi}(\mathcal{M}_k))^2 \right ] \\
    &=\mathbb{E}_{q(\pi|x)} \left [\sum_k \pi_k (x^T x - 2x^T \mathcal{D}_{\phi}(\mathcal{M}_k) +\mathcal{D}_{\phi}^T(\mathcal{M}_k) \mathcal{D}_{\phi}(\mathcal{M}_k)) \right ]\\
    &=x^T x - 2x^T\mathbb{E}_{q(\pi|x)} \left [ \sum_k \pi_k \mathcal{D}_{\phi}(\mathcal{M}_k) \right ] +\mathbb{E}_{q(\pi|x)} \left [ \sum_k \pi_k \mathcal{D}_{\phi}^T(\mathcal{M}_k) \mathcal{D}_{\phi}(\mathcal{M}_k)\right ]\\
    &=x^T x - 2x^T\sum_k \mathbb{E}_{q(\pi|x)}\left[\pi_k\right]\mathcal{D}_{\phi}(\mathcal{M}_k)+\mathbb{E}_{q(\pi|x)} \left [ \sum_k \pi_k \mathcal{D}_{\phi}^T(\mathcal{M}_k) \mathcal{D}_{\phi}(\mathcal{M}_k) \right ]\\
    &=x^T x - 2x^T\sum_k 
 \frac{\alpha^k_\theta(x)}{S} \mathcal{D}_{\phi}\left(\mathcal{M}_k\right)+\mathbb{E}_{q(\pi|x)} \left [ \sum_k \pi_k \mathcal{D}_{\phi}^T(\mathcal{M}_k) \mathcal{D}_{\phi}(\mathcal{M}_k) \right ]\\
    &=x^T x - 2x^T\sum_k 
 \frac{\alpha^k_\theta(x)}{S} \mathcal{D}_{\phi}\left(\mathcal{M}_k\right) + \sum_k \mathbb{E}_{q(\pi|x)}\left[\pi_k\right]\mathcal{D}_{\phi}^T(\mathcal{M}_k) \mathcal{D}_{\phi}(\mathcal{M}_k) \label{eq:b_15}\\
    &=x^T x - 2x^T\sum_k 
 \frac{\alpha^k_\theta(x)}{S} \mathcal{D}_{\phi}\left(\mathcal{M}_k\right) + \sum_k \frac{\alpha^k_\theta(x)}{S} \mathcal{D}_{\phi}^T(\mathcal{M}_k) \mathcal{D}_{\phi}(\mathcal{M}_k) \label{eq:b_16}\\
 &=\mathbb{E}_{z \sim \pi} \left [ (x - \mathcal{D}_{\phi}(\mathcal{M}_k))^T (x - \mathcal{D}_{\phi}(\mathcal{M}_k)) \right ]\\
    &=\mathbb{E}_{z \sim \pi} \left [ ||x - \mathcal{D}_{\phi}(\mathcal{M}_k)||_2^2 \right ].
\end{align}
We define $S = \sum_k \alpha_\theta^k(x) $, and $\pi = [\pi_1, \ldots, \pi_K]$. $\mathbb{E}_{q(\pi|x)}[\pi]$ in Equation~\ref{eq:b_15} of the ELBO derivation is equal to $\alpha_\theta^k(x)/S$ in Equation~\ref{eq:b_16} using the properties of the Dirichlet distribution. Therefore, $\pi_k = \alpha_\theta^k(x)/S$ holds. After we obtain the probabilities $\pi$s, we can get samples $z \sim Cat(z|\pi)$ using the Gumbel-Softmax trick that is explained in Section~\ref{sec:dvae}.

\section{Experimental Details}\label{appendix:experimental_details}

We use PyTorch framework in our implementation. We train all of the models for 150K iterations on all datasets, using a single NVIDIA A100 GPU. Our computation time for training EdVAE for 150K iterations varies within 12-21 hours based on the image size of the data, and training PixelSNAIL over the latents for 500 epochs is 9 hours. We use 128 as the batch size, and the Adam optimizer with an initial learning rate $1e^{-3}$, and follow the cosine annealing schedule to anneal the learning rate from $1e^{-3}$ to $1.25e^{-6}$ over the first 50K iterations. We rerun all of our experiments using the seed values of \textit{42}, \textit{1773}, and \textit{1}. 

We anneal the $\beta$ coefficient starting from 0 over the first 5K iterations with cosine annealing schedule. Based on the model and the dataset, the upper bound for the $\beta$ coefficient varies as the KL terms are different in different models, and the $\beta$ coefficient decides the reconstruction vs KL term tradeoff. We also follow the temperature annealing schedule $\tau = \text{exp}(-10^{-5}.t)$ for the Gumbel-Softmax where $\tau$ denotes the temperature, and $t$ denotes the global training step. We initialize the codebook embeddings of these models using a Gaussian normal distribution. In EdVAE, we clamp the encoder's output $z_e(x)$ to be maximum 20 before converting it to the $\alpha_\theta$ parameters for the training stability. We observe that after we obtain the training stability, we clamp too few variables which does not affect the integrity of the latent variables. We provide detailed analysis for the effects of logits clamping in Appendix~\ref{appendix:clamping}.

As datasets, we use CIFAR10, CelebA, and LSUN Church. CIFAR10 consists of 60,000 32 $\times$ 32 RGB images in 10 classes. Each class consists of the same amount of images. We use the default train/test split of the dataset. CelebA dataset consists of more than 200,000 celebrity images. We use the default train/val/test split of the dataset. As preprocessing, we perform center cropping of 140 $\times$ 140, and resize the cropped images to 64 $\times$ 64 using bilinear interpolation. LSUN Church consists of 126,000 256 $\times$ 256 RGB images of various churches. We use the default train/test split of the dataset. We resize the images to 128 $\times$ 128 resolution using bilinear interpolation to use.

For VQ-VAE-EMA and GS-VQ-VAE, we use the same architecture and hyperparameters as suggested in \cite{oord2017neural}. We set the $\beta$ coefficient for VQ-VAE-EMA's loss to 0.25, and the weight decay parameter for the EMA to 0.99. We initialize the codebook embeddings using a uniform distribution as in \cite{oord2017neural}. For GS-VQ-VAE, the upper bound for the annealed $\beta$ coefficient of the KL divergence is set to $5e^{-6}$ for all datasets. 

For SQ-VAE and VQ-STE++, we use the proposed architectures and hyperparameters, and replicate their results.

\begin{table}[t!]
\centering
    \caption{Notations of network layers used on all models.} 
    \small
    \begin{tabular}{ll}
    \bottomrule
        \textbf{Notation} & \textbf{Description}\\
        \hline
        $\mathrm{Conv}_{n}^{(7\times7)}$
        & 2D Conv layer (out\_ch$=n$, kernel$=7$, stride$=1$, padding$=3$) \\
        \hline
        $\mathrm{Conv}_{n}^{(4\times4)}$
        & 2D Conv layer (out\_ch$=n$, kernel$=4$, stride$=2$, padding$=1$) \\
        \hline
        $\mathrm{Conv}_{n}^{(3\times3)}$
        & 2D Conv layer (out\_ch$=n$, kernel$=3$, stride$=1$, padding$=1$) \\
        \hline
        $\mathrm{Conv}_{n}^{(1\times1)}$
        & 2D Convl layer (out\_ch$=n$, kernel$=1$, stride$=1$, padding$=1$) \\
        \hline
        $\mathrm{MaxPool}$
        & 2D Max pooling layer (kernel\_size$=2$)\\
        \hline
        $\mathrm{Upsample}$
        & 2D upsampling layer (scale\_factor$=2$)\\
        \hline
        $\mathrm{EncResBlock}_{n}$
        & $3 \times (\mathrm{ReLU}\to\mathrm{Conv}_{n}^{(3\times3)})\to\mathrm{ReLU}\to\mathrm{Conv}_{n}^{(1\times1)}$ $+$ identity \\
        \hline
        $\mathrm{DecResBlock}_{n}$
        & $\mathrm{ReLU}\to\mathrm{Conv}_{n}^{(1\times1)} \to 3 \times (\mathrm{ReLU}\to\mathrm{Conv}_{n}^{(3\times3)})$ $+$ identity \\
        \hline
    \end{tabular}
    \label{tab:building_blocks}
\end{table}

As dVAE is the baseline model for EdVAE, we describe the architecture of dVAE and EdVAE in detail. The common building blocks used in the encoders and the decoders are given in Table~\ref{tab:building_blocks}. For the following architectures, $w$ and $h$ denote the \textit{width} and the \textit{height} of the images. For CIFAR10 $w=h=32$, for CelebA $w=h=64$, and for LSUN Church $w=h=128$. We use a codebook $\mathcal{M} \in R^{512\times 16}$ for all of our experiments.

As the encoder of dVAE and EdVAE return a distribution over the codebook, the last dimensions of the encoders' outputs $z_e(x)$s are all equal to 512 for all datasets. After the quantization of $z_e(x)$s, the last dimensions of the decoders' inputs $z_q(x)$s are all equal to 16 for all datasets.

\textbf{Encoder:} $x\in\mathbb{R}^{w\times h\times3} \to \mathrm{Conv}_{n}^{(\text{kw} \times \text{kw} )} \to \left[\mathrm{EncResBlock}_{n}\right]_2 \to \mathrm{MaxPool} \to \left[\mathrm{EncResBlock}_{2*n}\right]_2 \to \mathrm{MaxPool} \to \left[\mathrm{EncResBlock}_{4*n}\right]_2 \to \mathrm{Conv}_{4*n}^{(1\times1)} \to z_e(x) \in \mathbb{R}^{\sfrac{w}{4}\times\sfrac{h}{4}\times 4*n}$

\textbf{Decoder:} $z_q(x)\in\mathbb{R}^{\sfrac{w}{4}\times\sfrac{h}{4}\times 16} \to \left[\mathrm{DecResBlock}_{4*n}\right]_2 \to \mathrm{UpSample} \to \left[\mathrm{DecResBlock}_{2*n}\right]_2 \to \mathrm{UpSample} \to \left[\mathrm{DecResBlock}_{n}\right]_2 \to \mathrm{ReLU}\to\mathrm{Conv}_{3}^{(1\times1)} \to \hat{x} \in \mathbb{R}^{w \times h \times 3}$

where $n$ is equal to 128 for all datasets, and $4*n$ is equal to the number of the codebook embeddings. \textit{kw} denotes the kernel size of the convolution layer. For CIFAR10 and CelebA $\text{kw}=3$, for LSUN Church $\text{kw}=7$.

Lastly, the upper bound for the annealed $\beta$ coefficient of the KL divergence in dVAE is set to $5e^{-5}$ for all datasets. On the other hand, we set the upper bound for the annealed $\beta$ coefficient of the KL divergence to $5e^{-7}$ for CIFAR10 while we use $1e^{-7}$ for the remaining datasets in EdVAE. We discuss the effects of the $\beta$ coefficient in  Appendix~\ref{appendix:beta_coeff}.

\section{Additional Experiments}\label{appendix:additional_results}

\subsection{Higher Temperature Usage with dVAE}\label{appendix:temp_exp}

We use temperature values of 2 and 5 in our current experiments, and observe that perplexity value obtained as 190 for CIFAR10 dataset slightly decreases to 170 and 180, respectively. Similarly, for CelebA dataset, perplexity value obtained as 255 decreases to 217 using temperature 2, and increases to 296 using temperature 5. The performance of dVAE is sensitive to temperature hyperparameter, and perplexity does not always increase with a high temperature. Therefore, using higher temperature is not an appropriate solution.

\subsection{Effects of Logits Clamping}\label{appendix:clamping}

To obtain the parameters of the Dirichlet distribution, $\alpha$s, we follow a common approach of logits clamping to stabilize the training since the exponential of logits might be really large. We conduct an ablation study to observe the effects of logits clamping, and present our findings in Table~\ref{tab:clamping}.

\begin{table}[t!]
\caption{Test perplexities using various max clamping values.}
\label{tab:clamping}
\centering
\begin{tabular}{llllll}
    \hline
    \multicolumn{1}{l}{\textbf{Max Clamping Value}} &\multicolumn{1}{l}{10} &\multicolumn{1}{l}{15} &\multicolumn{1}{l}{20} &\multicolumn{1}{l}{25} &\multicolumn{1}{l}{30}\\ 
    \hline
    \textbf{CIFAR10} &351 &421 &\textbf{425} &411 &1 \\
    \textbf{CelebA} &319 &376 &\textbf{386} &1 &1 \\
    \textbf{LSUN Church} &363 &375 &\textbf{393} &1 &1 \\
    \hline
\end{tabular}
\end{table}  

We observe that clamping the logits with smaller max values clamps some of the values in logits, and limits the range of positive values logits can have. This situation limits the representativeness of the logits, and leads to lower perplexities. On the other hand, using larger max values for clamping causes divergence in the training as the exponential of logits gets large, and the model cannot be trained. Therefore, the logits should be clamped eventually with proper values. If a proper max value can be selected, clamping acts as a regularizer at the beginning of the training, and the encoder naturally outputs logits with no values greater than the max clamping value after a few iterations. If the training is already stabilized, the max clamping value does not affect the performance dramatically as both 15 and 20 lead to similar results. Therefore, using 20 as the max value can be a mutual design choice.

\subsection{Effects of $\beta$ Coefficient}\label{appendix:beta_coeff}

\begin{figure}[t!]
    \centering
    \begin{subfigure}[b]{0.49\textwidth}
        \centering
        \includegraphics[width=0.9\textwidth]{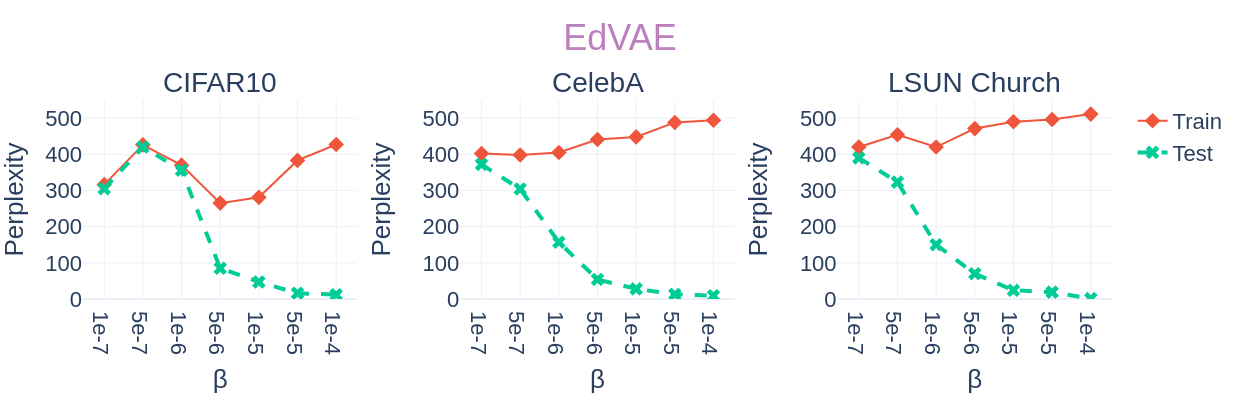}
        \caption{EdVAE.}
    \end{subfigure}
    \hfill
    \begin{subfigure}[b]{0.49\textwidth}
        \centering
        \includegraphics[width=0.9\textwidth]{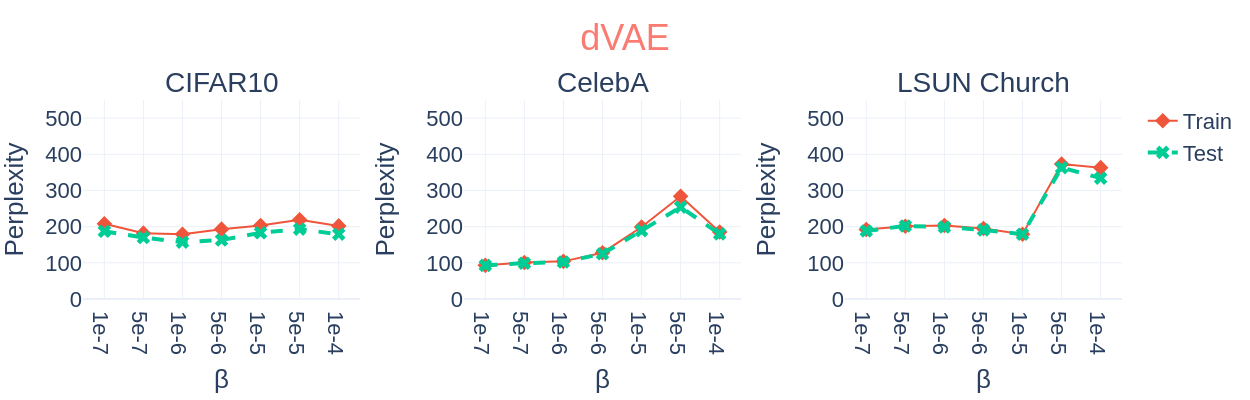}
        \caption{dVAE.}
    \end{subfigure}
    \caption{Effects of $\beta$ coefficient to the performance.}
    \label{fig:beta_coeff}
\end{figure}

We conduct additional experiments to observe $\beta$ coefficient's effects on the performance. We perform several experiments by changing the $\beta$ coefficient within [1e-7, 1e-4]. We repeat our experiments for dVAE and EdVAE using all of the datasets, and present our findings in Figure~\ref{fig:beta_coeff}.

We observe that our method is more sensitive to $\beta$ coefficient than dVAE, and EdVAE diverges when the $\beta$ coefficient increases. We think that the key factor to this sensitivity is the complexity introduced by the KL distance between our newly introduced posterior and prior, compared to the KL distance in dVAE. Therefore, fine-tuning $\beta$ emerges. Even though our original KL term brings some sensitivity to training and it requires a hyper-parameter tuning like most of the AI models, its contribution to the performance is non-negligible and essential.

Besides, the best performing $\beta$ coefficient for CIFAR10 dataset is slightly higher than the best performing $\beta$ coefficient of CelebA and LSUN Church datasets. Our intuition for this difference is that, reconstructing images with lower resolution as in CIFAR10 is less challenging than reconstructing images with higher resolution as in CelebA and LSUN Church. Therefore, increasing the $\beta$ coefficient from 1e-7 to 5e-7 improves the performance in CIFAR10 without hurting the reconstruction vs KL term tradeoff. On the other hand, 1e-7 to 5e-7 conversion slightly decreases the performance in CelebA and LSUN Church datasets since the reconstruction of the higher resolution images affects the reconstruction vs KL term tradeoff.

\bibliographystyle{elsarticle-num-names}
\bibliography{mybibfile}

\end{document}